\documentclass[11pt,a4paper]{article}
\usepackage[hyperref]{emnlp2020}
\usepackage{times}
\usepackage{latexsym}

\usepackage{graphicx}
\usepackage{framed}
\usepackage[normalem]{ulem}
\usepackage{booktabs}
\usepackage{array}
\usepackage{xspace}
\usepackage{comment}
\usepackage{enumitem}
\usepackage{amsmath}
\usepackage{tabularx}
\newcolumntype{L}{>{\raggedright\arraybackslash}X}
\usepackage[T1]{fontenc}
\usepackage[utf8]{inputenc}
\usepackage{caption}
\usepackage{subcaption}
\usepackage{amssymb}
\usepackage{amsmath}

\usepackage{microtype}

\aclfinalcopy % Uncomment this line for the final submission
\def\aclpaperid{1590} %  Enter the acl Paper ID here

\usepackage{array}
\newcommand{\PreserveBackslash}[1]{\let\temp=\\#1\let\\=\temp}
\newcolumntype{R}[1]{>{\PreserveBackslash\raggedleft}p{#1}}
\newcolumntype{C}[1]{>{\centering\arraybackslash} m{#1}}

\definecolor{purple}{rgb}{0.5,0,1}
\definecolor{dcyan}{rgb}{0.2,0.6,0.5}
\definecolor{light-gray}{gray}{0.95} % used in table

\definecolor{darkgreen}{RGB}{0,140,0}
\definecolor{darkred}{RGB}{200,0,0}
\definecolor{lightgreen}{RGB}{189,252,192}
\definecolor{lightred}{RGB}{255,205,212}
\definecolor{lightyellow}{RGB}{255,240,160}
\definecolor{lightblue}{RGB}{195,221,255}
\definecolor{lightpurple}{RGB}{232,209,255}

\newcommand{\redtext}[1]{\colorbox{lightred}{#1}\xspace}
\newcommand{\greentext}[1]{\colorbox{lightgreen}{#1}\xspace}

\newcommand{\bluetext}[1]{\colorbox{lightblue}{#1}\xspace}
\newcommand{\purpletext}[1]{\colorbox{lightpurple}{#1}\xspace}

\newcommand{\fbseries}{\unskip\setBold\aftergroup\unsetBold\aftergroup\ignorespaces}
\makeatletter
\newcommand{\setBoldness}[1]{\def\fake@bold{#1}}
\makeatother

\DeclareRobustCommand\sbseries{\fontseries{sb}\selectfont}
\DeclareTextFontCommand{\textsb}{\sbseries}

\newcommand{\unifiedQABig}{\textbf{\textsc{UnifiedQA}}\xspace}
\newcommand{\unifiedQA}{\textsc{UnifiedQA}\xspace}
\newcommand{\unifiedQABART}{\textsc{UnifiedQA}$_\text{BART}$}
\newcommand{\unifiedQATfive}{\textsc{UnifiedQA}$_\text{T5}$}

\newcommand{\script}[1]{{\small \texttt{#1}}}

\newcommand{\ignore}[1]{}
\newcommand{\ashish}[1]{{\small \color{blue} [AS: #1]}}

\newcommand{\daniel}[1]{#1}

\newcommand{\added}[1]{#1}

\title{
\vspace*{-0.5in}
{{\small \hfill EMNLP-Findings'20}\\
\vspace*{.25in}} 
\unifiedQABig: Crossing Format Boundaries with a Single QA System
}

\author{
Daniel Khashabi$^{1}$\ \ \;\; 
  Sewon Min$^{2}$\ \ \;\;   
  Tushar Khot$^{1}$\ \ \;\; 
  Ashish Sabharwal$^{1}$\ \ \;\;
  \\ 
  \textbf{Oyvind Tafjord}$^{1}$\ \ \;\; 
  \textbf{Peter Clark}$^{1}$\ \ \;\; 
  \textbf{Hannaneh Hajishirzi}$^{1,2}$ 
  \\
  \\
 $^1$Allen Institute for AI, Seattle, U.S.A.\\
 $^2$University of Washington, Seattle, U.S.A.  \\
}

\begin{document}

\maketitle

%%%%%%%%%%%%%%%%%%%%%%%%
\begin{abstract}
%\daniel{I like our previous title: ``Crossing Format Boundaries With a Single QA System'' (1) it clearly high-lights our main selling point (2) besides, the term ``unifiedqa'' also implicitly makes it clear that there is an artifact, without explicitly stating it; I don't like including ``pre-trained'' in the title; I feel that it over-shadows our key point (i.e., crossing format boundaries).}
%\daniel{Update: now among the 4 titles, I like the first and the last.}  \hanna{I don't feel strongly, but I like to have pre-trained in the title to show that people can take this pre-trained model and emphasize on it; Therefore, I vote for "A Pre-Trained QA Model Crossing Format Boundaries"  In the last title, unifiedQA refers to the process rather than the final system. }
Question answering (QA) tasks have been posed using a variety of formats, such as extractive span selection, multiple choice, etc. 
%, abstractive answer generation, multiple choice, and yes/no answers.
% \nick{Listing examples provides evidence for your point, but you could get more oomph with fewer words by citing a statistic, if possible. I.e. "To address different use-cases, researchers have proposed many QA formats. In ACL 2019, QA datasets spanned more than 15 (needs real number) distinct question types..."}. 
%This has led to format-specialized models, and even to an implicit division in the QA community based on format preferences and data availability. We argue that such boundaries are artificial and perhaps unnecessary over-complications. After all, the reasoning abilities we seek to teach and probe are not format specific. 
This has led to format-specialized models, and even to an implicit division in the QA community.  We argue that such boundaries are artificial and perhaps unnecessary, given the reasoning abilities we seek to teach are not governed by the format. 
As evidence, we use the latest advances in language modeling to build a \emph{single pre-trained QA model}, \unifiedQA, that performs  well across 20 QA datasets spanning 4 diverse formats. \unifiedQA performs on par with 8 different models that were trained on individual datasets themselves. Even when faced with 12 unseen datasets of observed formats, \unifiedQA performs surprisingly well, showing strong generalization from its out-of-format training data. Finally,  fine-tuning this pre-trained QA model into specialized models results in a new state of the art on 10 factoid and commonsense QA datasets, establishing \unifiedQA as a strong starting point for building QA systems.\footnote{\label{footnote:code}
% \hanna{instead of footnote, you can say: Code and demo are available at blah and blah}\daniel{On this stylistic point, i like the urls in footnote.}
% \url{https://unifiedqa.apps.allenai.org}
\url{https://github.com/allenai/unifiedqa}
%  \url{https://github.com/anonymous}
}
\end{abstract}

%%%%%%%%%%%%%%%%%%%%%%%%
\section{Introduction}

\begin{figure}[t]
    \centering
    \includegraphics[scale=0.69,trim=1.65cm 15.2cm 6.5cm 2cm, clip=true]{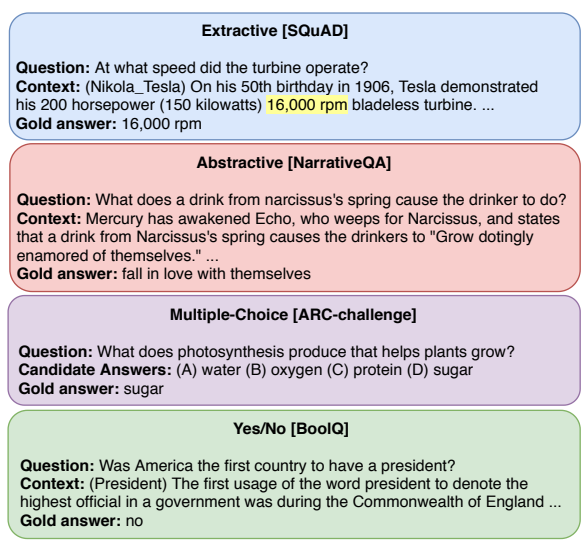}
    \caption{
        Four formats (color-coded throughout the paper) commonly used for posing questions and answering them: \bluetext{Extractive (EX)}, \redtext{Abstractive (AB)}, \purpletext{Multiple-Choice (MC)}, and \greentext{Yes/No (YN)}. Sample dataset names are shown in square brackets. We study generalization and transfer across these formats.
    }
    \label{fig:intro_figure}
\end{figure}

% \ashish{Daniel: did you say you can make \unifiedQA less bold? It's a little too strong in the abstract and some other places.}
Question answering is a common tool for assessing how well can computers understand language and reason with it. To this end, the NLP community has introduced several distinct datasets,
% including yes/no questions (e.g., BoolQ~\cite{clark-etal-2019-boolq}), extractive span selection (e.g., SQuAD~\cite{rajpurkar-etal-2016-squad}), multiple-choice (e.g., MCTest~\cite{richardson-etal-2013-mctest}), and abstractive questions (e.g., NarrativeQA~\cite{kocisky-etal-2018-narrativeqa}).
with four popular \emph{QA formats} illustrated in Fig.~\ref{fig:intro_figure}.
%These formats differ not only in how the question is presented but also in some implicit assumptions.
For instance, some datasets expect the answer to be  ``yes'' or ``no'', or a unique answer span in the associated paragraph (as opposed to multiple or no spans). 
These differences have motivated their study in silos,
% isolated from each other
% \nick{I'm not sure if this claim is entirely true. A number of datasets mix different formats, like DROP for example. Similarly, I think you'll have to acknowledge closely related work (i.e., https://arxiv.org/pdf/1806.08730.pdf) pretty early on in this introduction.}.
often encoding QA format  into the model architecture itself. Efforts to exploit multiple datasets remain largely restricted to a single format. For example, \citet{clark2019f} limit consideration to multiple-choice datasets, while \citet{talmor-berant-2019-multiqa} focus their generalization study on extractive span prediction models.
To the best of our knowledge, no single QA system targets, not to mention excels at, all of these formats.
% \nick{I think T5 and The Natural Language Decathlon papers could reasonably claim to be such systems}.

\begin{figure*}[t]
    \centering
    \includegraphics[scale=0.66,trim=10.1cm 16.5cm 10cm 2.5cm]{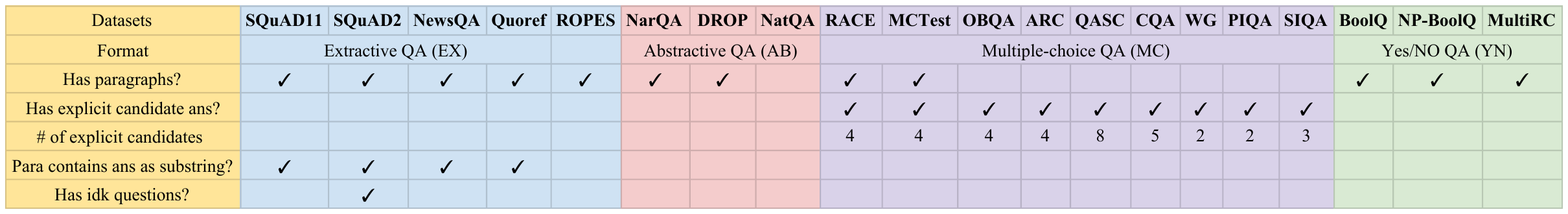}
    \caption{Properties of various QA datasets included in this study: 5 extractive (EX), 3 abstractive (AB), 9 multiple-choice (MC), and 3 yes/no (YN). `idk' denotes `I don't know' or unanswerable questions. 
    % Regents denotes both 4th and 8th grade datasets. 
    BoolQ represents both the original dataset and its \emph{contrast-sets} extension BoolQ-CS; similarly for ROPES, Quoref, and DROP.
    % \ashish{format is stated twice. Replace the 2nd row with what's currently the last row}
    %\sewon{Looks like NQA is used instead of NarQA, and NatQA is missing.}
    %\daniel{updated}
}
    \label{fig:datasets:properties}
\end{figure*}

This raises the question: \emph{Can QA models learn linguistic reasoning abilities that generalize across formats?}
Our intuition is simple: while question format and relevant knowledge may vary across QA datasets, the underlying linguistic understanding and reasoning abilities are largely common. A multiple-choice model may, therefore, benefit from training on an extractive answers dataset.
Building upon this intuition, we present a single pre-trained QA system, named \unifiedQA, that exploits information across 4 different QA formats to achieve  strong performance across 20 different
factoid and commonsense QA datasets listed in Fig.~\ref{fig:datasets:properties}.

In this work, we advocate for a unifying view of QA formats by building a format-agnostic QA system.
Our work leverages recent progress in text-to-text pre-trained neural models, specifically T5~\cite{raffel2019exploring} and BART~\cite{lewis2019bart}, but with a strong focus on differing QA formats. 
This paradigm allows unifying many NLP models, which formerly had task-specific designs, into a single text-to-text framework. 
% Most approaches fine-tuned a different set of parameters for each end task \cite{raffel2019exploring,radford2019language}, and when
% they have attempted to make a single model for
% different NLP tasks \cite{raffel2019exploring}, they
% have \underline{underperformed} compared to the standard pretraining plus fine-tuning setup, with a need for explicit task-specific prefixes
% \begin{comment}
Previous work uses textual prefixes to explicitly define 
the task associated with each input instance~\cite{raffel2019exploring,radford2019language};
often such attempts to build a single model for
multiple NLP tasks \underline{underperform} the standard pre-training plus fine-tuning setup (a model per task)~\cite{raffel2019exploring}.
% \tushar{This seems to conflate two different setups which might be confusing. So T5 folks tried two things (1) task-specific prefix but one set of parameters: doesn't do too well and worse than standard pre-training + fine-tuning. (2) fine-tune a different set of parameters for each end task and that is basically the same as `` standard pre-training plus fine-tuning setup''. If this is correct, we should drop ``and fine-tunes a different set of parameters for each end task''.} 

Our work narrows down the scope to tasks that stay within the boundaries of QA,  demonstrating that a unified text-to-text paradigm can, in fact, be successful across different QA tasks and formats. 
% \hanna{ We need one sentence explaining how we did it: }
We develop a single pre-trained QA model by training text-to-text models on a set of seed QA datasets of multiple formats, taking natural text as input, without using format-specific prefixes. Our experiments show that \unifiedQA  can be applied as-is to different QA tasks, %, takes in natural text inputs without explicitly specifying  a task-specific prefix, 
generalizes well to other unseen datasets (zero-shot), and with further fine-tuning  achieves state-of-the-art results on many QA tasks including commonsense and factual datasets.  

\paragraph{Contributions.}
This work advocates for a unified view of different QA formats, and for building format-agnostic QA systems. To support this view, we present \unifiedQA, a single pre-trained QA system that works well on and generalizes to datasets with different formats (\S\ref{subsec:generalization}), while performing on par with state-of-the-art dedicated systems tailored to each dataset (\S\ref{subsec:union:vs:single:dataset}).
% \daniel{\sout{it is trained on}}.
Additionally, fine-tuning  \unifiedQA into specialized systems sets a new state of the art for 10 datasets 
% \ashish{we should clarify somewhere soon how we count; a glance at that table suggests it's SOTA on 9 datasets; same for other numbers like 17, etc.} 
(\S\ref{subsec:sota}), establishing it as a powerful starting point for QA research. Our findings demonstrate that crossing QA format boundaries is not only qualitatively desirable but also quantitatively beneficial.
% \daniel{I added references to the corresponding sections  make it easier for people to trace the contributions. }

%%%%%%%%%%%%%%%%%%%%%%%
\section{Related Work}

Several QA efforts have studied generalization across datasets of a \emph{single} format. For instance, in MultiQA, \citet{talmor-berant-2019-multiqa} study generalization and transfer, but only across extractive span selection datasets. Further, while they show strong leave-one-out style results, they find a single system performs substantially worse than one tuned to each dataset. In ORB, \citet{dua2019orb} propose a multi-dataset evaluation benchmark spanning extractive and abstractive formats.
% They encourage consideration of a broad combination of datasets, but
However, that study is limited to an \emph{evaluation} of systems, falling short of addressing how to build such generalized models.
% \tushar{Don't they also have datasets spanning formats e.g. NarrativeQA. In theory, ORB could add BoolQ and RACE to their test sets. One key difference to point out: These are finally \emph{test} sets. They do not consider how to build generalizable models or answer the question: Can different formats help with building better models?}\ashish{good point! will help differentiate from ORB, which we have got to cite.}
The MRQA shared task~\cite{fisch-etal-2019-mrqa} focuses on span-prediction datasets. Unlike all these efforts, our goal is to investigate transfer and generalization across different QA formats, as well as to build a single system that does this well.

Exploiting commonality across machine learning tasks has a rich history studied under transfer learning~\cite{caruana1997multitask,clark2019bam}. 
\citet{mccann2018natural} and \citet{keskar2019unifying} study transfer among various NLP tasks by casting them into a single QA format---an elegant transfer learning approach but orthogonal to the goal of this work.
As noted earlier, \citet{raffel2019exploring} investigate the transfer between several diverse NLP tasks (machine translation, summarization, etc). Their key contribution is a text-to-text framework, and a powerful model called T5, that makes it easier to mix multiple tasks by encoding both inputs and outputs as text. 
% To define the inputs of their diverse task, they employ 
% The work mixes a diverse set of tasks, by defining tasks with 
They rely on textual prefixes to explicitly define the task corresponding to each input instance.
% a \emph{task-specific} textual prefix for each input sequence to help their model differentiate between tasks.
While we build upon their framework, 
% In this work, we build upon their text-to-text framework and model \nick{I'd say the text-to-text framework is separate from the T5 model.}, T5. However, 
we narrow our focus to variations of QA. 
This allows us to achieve strong results while avoiding reliance on any format-specific prefixes. 
Our models \emph{learn to infer} the format of each input question based on its content (e.g., whether the phrasing of the question demands a yes/no answer).
% \ashish{good! may be drop `based on their statement' or use something like `learn to infer the QA format from the input'}\daniel{hb now?}
Moreover, we are able to demonstrate generalization across QA tasks, which prior work failed to achieve presumably due to its focus on too broad a set of NLP tasks. 

% \daniel{TODO: somewhere cite: NL-Decathlon paper: \cite{mccann2018natural}}

%%%%%%%%%%%%%%%%%%%%%%%%

% \hanna{It is better to get to UnifiedQA sooner. The reader is all anxious to learn about it. I vote to move multi-format training section here and  rename this section to: UnifiedQA: Multi-format Training; Then add a sub-section to this (call it UnifiedQA and explain fine-tuning in it). We don't need to know all the details about the datasets to understand multi-format training, we just need a paragraph reminding different formats and how you do the multi-format training. Then, you can define the datasets after UnifiedQA section; And then list all the experiments. }
% \ashish{Agreed. We can start the `UnifiedQA: Multi-format Training' section with what you have as the first sentence of each of EX, AB, MC, and YN paragraphs in the Datasets section. Could be a bulleted list, also introducing the 2-letter acronyms. The current `3.2 Evaluation Metrics' subsection should then move as a subsection right after `Text-to-Text Encoding', and then `UnifiedQA' as the 3rd subsection.}

%%%%%%%%%%%%%%%%%%%%%%%%%%%%%%%%%
\section{\unifiedQABig: Multi-format Training}
\label{sec:multiformat-training}

% \ashish{Turned this into a abstract description.  The more we can do this and keep the description less tied to specific choices, the more we get 3 benefits: (1) it brings out the methodological contribution; (2) if someone disagrees with a specific choice we make, they still won't dismiss the whole work; (3) makes it clear that \unifiedQA model can be easily extended and isn't frozen.}
% \daniel{I like the abstraction!}

Suppose we would like to train a unified QA model that can operate over $k$ formats $F_1, F_2, \ldots, F_k$. For each format $F_i$, suppose we have $\ell_i$ datasets sets $D^i_1, D^i_2, \ldots, D^i_{\ell_i}$ where $D^i_j = (T^i_j, E^i_j)$ includes both training and evaluation examples. In some cases, the training set $T^i_j$ may be empty or we may want to ignore it in order to treat $D^i_j$ as an `unseen', \emph{evaluation-only} 
% \daniel{\emph{evaluation-only}?}
% \ashish{how about now? we refer to unseen here and there, so mentioning here seems useful}\daniel{looks good!}
dataset and assess a model's generalization to it.

We use the text-to-text paradigm to convert each training question $q$ in format $F_i$ into a \emph{plain-text} input representation $\mathit{enc}_i(q)$. This conversion uses a natural encoding process that will be described shortly (\S\ref{sec:datasets:encoding}) for four common QA formats, and is easily extensible to other formats as well. We follow a simple approach of creating a mixed training pool consisting of all available training instances: 
$$
  \tilde{T} = \bigcup_{i=1}^k \bigcup_{j=1}^{\ell_i} \Big\{ \mathit{enc}_i(q) \mid q \in T^i_j \Big\}
$$
% \ashish{double-check}\tushar{$r_i$ since the transformation is format-specific. Also we can then mention that for any new format, we just need to define $r_i$} 
% The pooled training data, $\tilde{T}$, is shuffled uniformly at random and training batches are then drawn from it such that
% \ashish{updated the following to reflect the change}
Training batches are drawn from this pooled data, $\tilde{T}$, by including each $q \in T^i_j$ with a probability proportional $1 / |T^i_j|$.
Each batch thus, on average, contains the same number of instances from each training set, regardless of its size. 
Similar treatments of task mixing have also been adopted by~\citet{arivazhagan2019massively} and \citet{raffel2019exploring}. 
% a mixture of instances that aligns with the original proportion of training set sizes and as well as the question format they represent. 
% \daniel{dropped an email to T5 authors to make sure that this is accurate}
% \daniel{update: per our correspondence last night, batches have a uniforming mixing of different datasets. }
As our experiments will show, our multi-format mixing approach works  well. It clearly highlights the value of training on out-of-format data and confirms our intuition that there are strong ties across QA formats in terms of the underlying reasoning abilities.\footnote{A more sophisticated teaching curriculum~\cite{sachan2016easy} or approaches such as model distillation and teacher annealing~\cite{clark2019bam} are likely to further improve the performance of the resulting unified model, bolstering the strength of our advocacy for a unified view of all QA formats. We leave their exploration to future work.}

\begin{comment}
The experiments have the following high-level organization: The first part \S\ref{subsec:pair} mixes \emph{pairs} of formats. The second part  of the work \S\ref{subsec:union} uses the lessons we learned in the first part (like what formats/datasets to mix, etc) to build a a single system that does well across multiple formats. 
% , i.e. the long-promised UniversalQA (T6?).  
\end{comment}

Our unified question-answering system is based on the recent text-to-text frameworks, particularly, T5~\cite{raffel2019exploring} and BART~\cite{lewis2019bart}.
We first define a unifying encoding of the instances across various formats (\S\ref{sec:datasets:encoding}). We then introduce \unifiedQA (\S\ref{subsec:unifiedQA}) that is a QA system trained on datasets in multiple formats, indicating new state-of-the-art results on 10 datasets and generalization to unseen datasets.

\begin{comment} \daniel{should this pragraph be titled ``Roadmap.''?}
\ashish{sure, in which case, move it to the end of the intro and update it to include section 3 as well}
\daniel{If we name it ``Roadmap of experiments.'' it could stay here?}
\ashish{it would be a very unusual place to talk about experiments. best to move it away}
We first define a unifying encoding of the instances across various formats (in \S\ref{sec:datasets:encoding}).
Then in \S\ref{subsec:pair} we evaluate the impact of training on out-of-format datasets, e.g., does training our model on a mix of MC and EX dataset leads to improved scores on EX tasks? As we empirically show, combining datasets from two different formats does lead to improved performance. 
This motivates the design of \unifiedQA, described in \S\ref{subsec:unifiedQA}). We finally perform extensive evaluations on \unifiedQA that show new state-of-the-art results on 7 datasets and generalization to unseen datasets.
\end{comment}

%%%%%%%%%%%%%%%%%
\subsection{Text-to-Text Encoding}
\label{sec:datasets:encoding}

We convert each of our target datasets into a text-in/text-out format~\cite{raffel2019exploring,lewis2019bart,radford2019language}.
% \ashish{should cite T5; they were the first. then clarify the difference (no task specific prefix)} \daniel{done in the intro and related work}
The question always comes first, followed by some additional information (context paragraph or candidate answers, or both). We use ``\script{\textbackslash n}'' separators between different parts of the input. This ensures having a human-like encoding while not making it overly-specific to a certain format. %Below are several examples. Here are several ways the encoding is done: 

Our unified model incorporates the following four common question-answering formats. Specific   datasets within them are deferred to Section~\ref{sec:datasets}.
\begin{description}[topsep=1pt,itemsep=1pt,leftmargin=0pt]
    \item [Extractive (EX)]  questions $Q$ include a context paragraph $C$ (typically a paragraph) and require models to extract the answer as a substring from the context. In some datasets, `unanswerable' can sometimes be the correct response.
    
    \item [Abstractive (AB)]  questions $Q$ require models to produce answers that are often not mere substrings of the provided context paragraph $C$.
    
    \item [Multiple-choice (MC)]  questions $Q$ come with a set of candidate answers $\{A_i\}$, of which generally exactly one is correct. In some cases, they also include a context paragraph $C$.
    
    \item [Yes/No (YN)]  questions $Q$ expect a `yes' or `no' answer as the response and may include a context paragraph $C$.
\end{description}

%\daniel{
 %   FYI: From here on I always use the combination ``context paragraph''; I dislike using ``context'' alone (with no ``paragraph'' next to it) since it trivializes ``context'', which is  is a much broader concept that a ``paragraph''. 
%}

% \ashish{TODO: The following is described by examples, which isn't ideal. Change to first have a clear, semi-formal specification of the encoding for each format. Then go into examples. Two possibilities: (1) have all encodings in a single table, at the specification level of \script{question \textbackslash n candidates}; such a table would be easy for readers to refer to; (2) merge such specification in the bulleted list above.}
% \daniel{reworked this section; feel free to moved it to where it belongs (cf. the previous sentence):}

Table~\ref{tab:example:encodings} provides examples of the natural input and output encoding for each of these formats, where both input and output representations are raw text. There is no explicit information regarding a question being an MC question or having exactly four candidate answers. Specifically, MC questions without any context paragraph are encoded as 
\script{question \textbackslash n (A) c1 (B) c2 $\hdots$} where \script{c1}, \script{c1}, $\hdots$ are the set of candidate answers (see the example from ARC dataset).
% \daniel{it should imply that (?): For consistency we always use \script{(A)}, \script{(B)}, $\hdots$ to indicate the candidates. Even if the original question comes with (i) (ii), ... we convert them to our our format.}  
% \ashish{yes, it does!}
% \ashish{can \script{candidates} be a bit more formal? To state whether it is \script{c1 | c2 | c3 ...} or \script{(A) c1 (B) c2 ...}. The example hints at it but doesn't quite say what would happen if the original question had (i) (ii) (iii) etc for candidate numbering. Overall, let's be more precise.} 
If the question includes a context paragraph, it is appended after the candidate answers: \script{question  \textbackslash n (A) c1 (B) c2 $\hdots$ \textbackslash n paragraph}, as shown in the example from the MCTest dataset. Questions in the other three formats (EX, AB, and YN) are encoded simply as \script{question \textbackslash n paragraph}. 

To re-emphasize, unlike prior work \cite{raffel2019exploring}, we do not specify any task-, dataset-, or format-specific prefixes in the input representation. Whether the answer should be extracted or abstracted, and whether from the provided context paragraph or candidate answers (or the fact that these even are candidate answers) is expected to be inferred by the system.

\begin{table}[t]
    \centering
    \includegraphics[scale=0.65,trim=8cm 6.3cm 0cm 2cm]{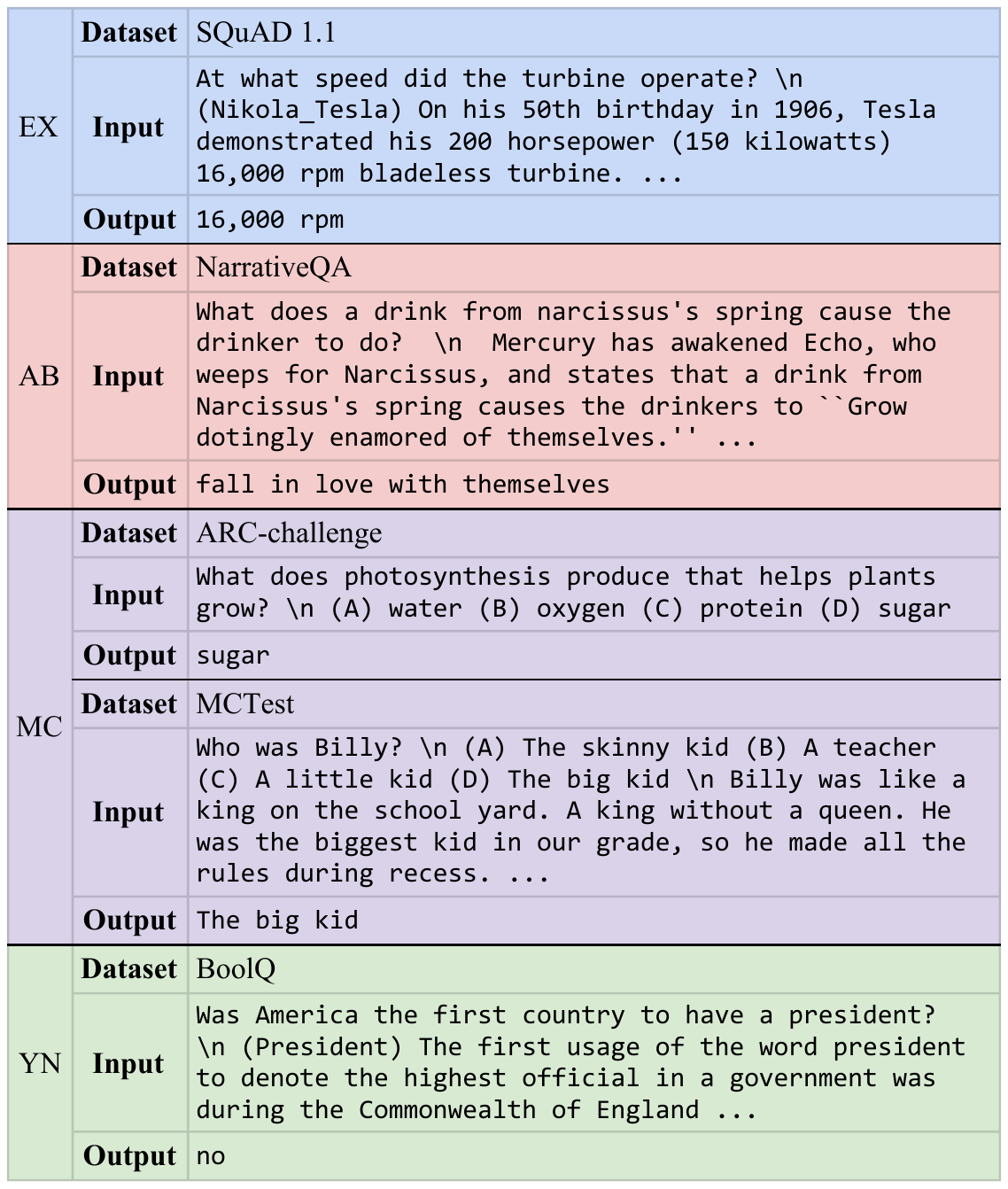}
    \caption{Example text-to-text encoding of instances.}
    \label{tab:example:encodings}
\end{table}

\subsection{\unifiedQABig: The Pre-Trained Model}
\label{subsec:unifiedQA}

The specific pre-trained QA model we provide and use in all our experiments is trained on representative datasets for each of the 4 formats discussed earlier. We empirically chose the following 8 \emph{seed datasets} for training \unifiedQA,\footnote{Future references to `\emph{seed dataset}' point to the QA datasets used in this section.}
based on their effectiveness in our pilot study (details deferred to Section~\ref{subsec:pair}) assessing which datasets are most valuable for out-of-format training:
%\tushar{Didn't get the part following ``of which out-of-format...''. }\ashish{better?}
% \hanna {let's call these datasets as seed datasets; it is easier to refer to them later. }
\begin{itemize}[noitemsep]
    \item EX: SQuAD 1.1, SQuAD 2.0
    \item AB: NarrativeQA
    \item MC: RACE, ARC, OBQA, MCTest
    \item YN: BoolQ
\end{itemize}

One can easily use other combinations of formats and datsets to create variants of our \unifiedQA model, or extend it as future datasets become available or new formats are introduced.

Unless otherwise noted, we use the largest available T5 model (11B parameters) as the starting point for training our model and call the system \unifiedQA. We also report results of training our system with BART$_\text{large}$, referred to as  \unifiedQABART\ (see \S\ref{subsec:sota}). 
%The majority of the experiments are done on T5, except \S\ref{subsec:sota} where we show the results of experiments with BART as well. 
Details on the parameters of the models used are deferred to Appendix~\ref{appendix:subsec:hyperparams}. 
Similar to pre-trained language models, the resulting pre-trained QA model can be used as a starting point for fine-tuning on other QA datasets.

%\tushar{Seems a bit out-of-place to talk about experimental results.}\ashish{better? see above}
% We empirically show that fine-tuning \unifiedQA achieves better performance as opposed to a vanilla language model. 

%%%%%%%%%%%%%%%%%%%%%%%%%%%%%%%%%%%%%%%
\section{Formats and Datasets}
\subsection{Datasets}
\label{sec:datasets}
% \hanna{I suggest to distinguish between "seed datasets" and "evaluated datasets"}

% \hanna{if no space, this section can be very much summarized.}
% \ashish{+1. we are spending 1 page on 4.1. Let's move this as-is to the appendix, and have a really succinct version here. Cover dataset names + citations, and only point out unusual / important pieces of info (e.g., that DROP requires math skills (worth pointing out as it's different)).}
We evaluate \unifiedQA on 20 existing datasets that target different formats as well as various complex linguistic phenomena. 
Fig.~\ref{fig:datasets:properties} summarizes key properties of our datasets
%(whether it comes with a paragraph, 
% domain of the questions/paragraphs,
%whether the paragraph explicitly contains the answer, whether there are candidate-answers as part of the input, etc).
(whether it comes with a paragraph or answer candidates, whether the paragraph explicitly contains the answer, etc).
Most importantly, they are grouped into several formats/categories as described below. 
% : \textbf{EX} (extractive questions), \textbf{MC} (multiple-choice), \textbf{YN} (yes-no questions), \textbf{AB} (abstractive questions). 
% \nick{I don't think you need to list the categories here, you can just say "described below" or something and let them read the paragraphs.}
% From each format, I have 1 or 2 datasets selected to train on. For example, in the YN category, I am training on BoolQ. 
Table~\ref{tab:statitstics} gives certain statistics of these datasets. We next provide a summary enumerating these datasets, with additional details deferred to Appendix~\ref{appendix:sec:datasets}. 
%\sewon{A bit worried that IR versions (on ARC-easy, ARC-chel, QASC and NatQA) are not mentioned in this section, although it seems to be an important detail. How about adding a brief sentence, including how docs/passages are retrieved for those settings?}
%\daniel{We talk abt `w/ IR' in \S\ref{subsec:sota} since that's the only place this notation shows up. See if the relevant description is still not clear? }
% \sewon{Also, we did retrieval for NatQA but we did not put `w/ IR'. Putting `w/ IR', or otherwise make it clear that we did retrieval might be good.} \daniel{done.}
%\sewon{Below, each paragraph begins with the requirement for each format but they seems a bit repetitive to paragraphs in Section 3.1. It seems to be fine to remove them and just list datasets if we run out of space.}

\paragraph{Extractive QA (EX).}% All the datasets in this format require models to extract the answer to a given question as a substring from a context paragraph. 
Among the datasets in this popular format, we adopt SQuAD 1.1~\cite{rajpurkar-etal-2016-squad}, SQuAD 2~\cite{rajpurkar-etal-2018-know}, NewsQA~\cite{trischler-etal-2017-newsqa}, Quoref~\cite{dasigi-etal-2019-quoref}, ROPES~\cite{lin-etal-2019-reasoning}.

\paragraph{Abstractive QA (AB).}
%All the datasets in this format require models to produce answers that are often not mere substrings of the given context paragraph. 
The datasets used from this format are: 
NarrativeQA/NarQA~\cite{kocisky-etal-2018-narrativeqa}, 
the open-domain version of NaturalQuestions/NatQA~\cite{kwiatkowski-etal-2019-natural}, and
DROP~\cite{dua-etal-2019-drop}. 
%\sewon{I think NatQA is usually considered as extractive (the answer is always a span from Wikipedia), although it would depend on the definition of extractive.}
%\daniel{Yeah this can be labeled either EX or AB; I decided to go with AB because there is no guarantee the answer would explicitly be found in the retrieved paragraph, I believe. I don't have a strong opinion on this.  }
%\sewon{Makes sense!}

\begin{table}[t]
    \centering
    \small
    \resizebox{\linewidth}{!}{
    \begin{tabular}{lC{0.9cm}C{0.9cm}C{1.3cm}C{0.9cm}C{0.7cm}C{0.7cm}}
        \toprule
        Dataset & Train set size & Eval. set size & Best published & 95\% CI (\%) & Input length & Output length  \\
        \midrule
        SQuAD 1.1 & 87$k$ & 10$k$ & 95.6 & 0.4 & 136.2 & 3.0 \\ 
        SQuAD 2.0 & 130$k$ & 11$k$ & 91.2 & 0.5 & 139.9 & 2.6 \\ 
        NewsQA & 76$k$ & 4$k$ & 66.8 & 1.4 & 606.6 & 4.0 \\ 
        Quoref & 22$k$ & 2$k$ & 86.1 & 1.5 & 352.7 & 1.7 \\ 
        Quoref-CS & - & 700 & 55.4 & 3.6 & 324.1 & 2.2 \\ 
        ROPES & 10$k$ & 1.4$k$ & 61.1 & 2.5 & 169.1 & 1.4 \\ 
        ROPES-CS & - & 974 & 32.5 & 3.0 & 182.7 & 1.3 \\ 
        \midrule
        NarQA & 65$k$ & 21$k$ & 58.9 & 0.7 & 563.6 & 6.2 \\ 
        NatQA & 79$k$ & 3.6$k$ & 42.2 & 1.6 & 607.0 & 2.2 \\ 
        DROP & 77$k$ & 9$k$ & 89.1 & 0.6 & 189.1 & 1.6 \\ 
        DROP-CS & - & 947 & 54.2 & 3.2 & 206.0 & 2.1 \\ 
        \midrule
        RACE & 87$k$ & 4$k$ & 89.5 & 0.9 & 317.9 & 6.9 \\ 
        OBQA & 4$k$ & 501 & 80.0 & 3.3 & 28.7 & 3.6 \\ 
        MCTest & 1.4$k$ & 320 & 86.5 & 3.4 & 245.4 & 4.0 \\ 
        ARC (easy) & 2$k$ & 2$k$ & 80.0 & 1.7 & 39.4 & 3.7 \\ 
        ARC (chal.) & 1$k$ & 1$k$ & 67.8 & 2.9 & 47.4 & 5.0 \\ 
        % Regents & 1$k$ & 1$k$ & 3.1 & 51.0 & 4.9 \\ 
        CQA & 9.7$k$ & 1.2$k$ & 79.1 & 2.2 & 26.8 & 1.5 \\ 
        WG  & 40.3$k$ & 1.7$k$ & 67.5 & 2.2 & 25.2 & 3.0 \\ 
        PIQA & 16.1$k$ & 3$k$ & 79.4 & 1.4 & 49.6 & 20.2 \\ 
        SIQA & 33.4$k$ & 2.2$k$ & 78.0 & 1.7 & 37.3 & 4.7 \\ 
        \midrule
        BoolQ & 9$k$ & 3$k$ & 91.0 & 1.0 & 105.1 & 1.0  \\ 
        BoolQ-CS & - & 461 & 71.1 & 4.0 & 108.9 & 1.0  \\ 
        NP-BoolQ & 10$k$ & 3$k$ & 78.4 & 1.4 & 106.2 & 1.0  \\ 
        MultiRC & - & 312 & 91.7 & 2.6 & 293.3 & 1.0 \\ 
        \bottomrule
    \end{tabular}
    }
    \caption{
        Dataset Statistics. CQA, OBQA, WG, and NarQA refer to CommonsenseQA, OpenBookQA, Winogrande, and NarrativeQA, respectively. The CI column shows the \added{upper 95\% confidence interval for the evaluation set as a percentage, based on the Wilson test around the mean score listed as a percentage in the best known performance column.} Input and output representation lengths are measured in the number of tokens and averaged across the dataset.
        % \ashish{Added a column for 95\% Confidence Interval using \url{http://www.quantitativeskills.com/sisa/statistics/onemean.htm} with mean as above}
        % \sewon{Marginal comment: NQA is not very intuitive because we have three datasets that can be NQA (NewsQA, NarrativeQA, Natural Questions), and other papers haven't really called NarrativeQA as NQA. How about using the other name, like NarQA or NarrativeQ? (I think NQ is fine as other papers have been calling it NQ.)
        % }\ashish{+1 to NarQA or NrQA}
        %\sewon{What is CI? (i.e. CI for what?)}
        %\sewon{It's just my personal opinion but not sure if CI \& input/output length are important.. Just in case you need to shorten!}
        %\daniel{
        %    informally speaking, it is a helpful measure to see whether the improvement on each dataset is meaningful. 
        %    technically, a CI indicates whether the margin of difference between two accuracy values for which one cannot *reliably* reject the null-hypothesis (here, two systems having the same accuracy). 
        %}
        % \sewon{I updated NatQA statistics.
        % Prev. NatQA Statistics were in RC setup.
        % Does `Eval' mean dev or test? For now, I put the test. (79k, 8.8k, 3.6k for the train, dev, test.)  }. % DK: it's for eval set. Thanks! 
        }
    \label{tab:statitstics}
\end{table}

\paragraph{Multiple-choice QA (MC).}
%All the datasets in this format contain questions that come with candidate answers. 
We use the following MC datasets: 
MCTest~\cite{richardson-etal-2013-mctest}, RACE~\cite{lai-etal-2017-race}, OpenBookQA/OBQA~\cite{mihaylov-etal-2018-suit}, ARC~\cite{clark2018think,clark2016combining}, QASC~\cite{khot2019qasc}, CommonsenseQA/CQA~\cite{talmor-etal-2019-commonsenseqa}, PIQA~\cite{bisk2019piqa}, SIQA~\cite{sap2019socialiqa}, and Winogrande~\cite{sakaguchi2019winogrande}.  Several of the MC datasets do not come with accompanying paragraphs (such as ARC, QASC, OBQA). For most of this the work, we keep the questions as is with no additional retrieval (unless otherwise mentioned). %, except in \S\ref{subsec:sota} where we use IR to get numbers comparable to earlier work. 
One other variability among these datasets is their number of candidate answers. While many datasets have four candidates (see Fig.~\ref{fig:datasets:properties}), others have more. Later (in \S\ref{subsec:generalization}) we will see that our approach generalizes to datasets with different numbers of candidates, even if such questions have not been seen during training.

\paragraph{Yes/No QA (YN).}
%All the datasets in this format contain questions that could be responded with yes/no answers. One can think of these as multiple-choice questions with 2 candidates; however, they're usually treated differently. 
The YN datasets we use are BoolQ~\cite{clark-etal-2019-boolq} and a naturally-perturbed version of this dataset, BoolQ-NP~\cite{khashabi2020naturalperturbations}, and the binary (yes/no) subset of MultiRC~\cite{khashabi-etal-2018-looking}. 

\paragraph{Contrast-sets.}
Additionally, we use \emph{contrast-sets}~\cite{gardner2020evaluating} for several of our datasets (denoted with ``CS''): BoolQ-CS, ROPES-CS, Quoref-CS, DROP-CS. 
These evaluation sets are expert-generated perturbations that deviate from the patterns common in the original dataset.

 \begin{table*}[t]
     \centering
    \includegraphics[scale=0.64,trim=7.5cm 17.6cm 9.0cm 2cm, clip=false]{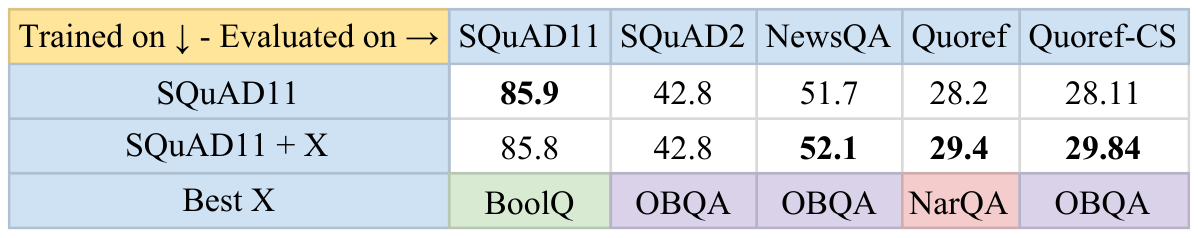}
    \includegraphics[scale=0.68,trim=7.5cm 17.6cm 8.1cm 2cm, clip=false]{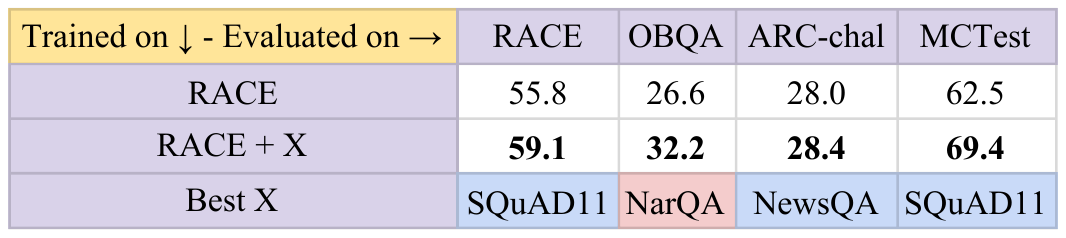}
    \includegraphics[scale=0.70,trim=8.7cm 17.6cm 7.1cm 1.5cm, clip=false]{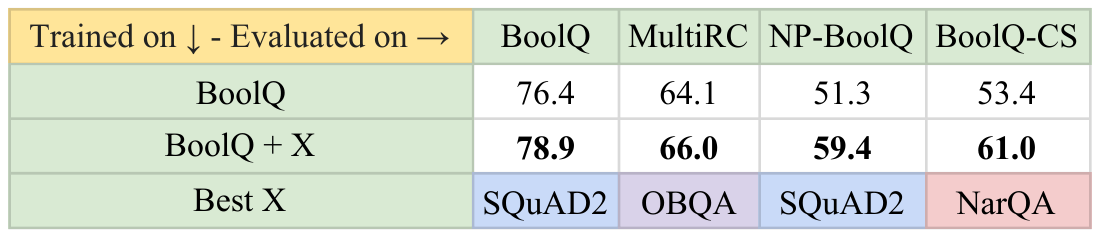} 
     \includegraphics[scale=0.70,trim=10cm 17.6cm 9.1cm 1.5cm, clip=false]{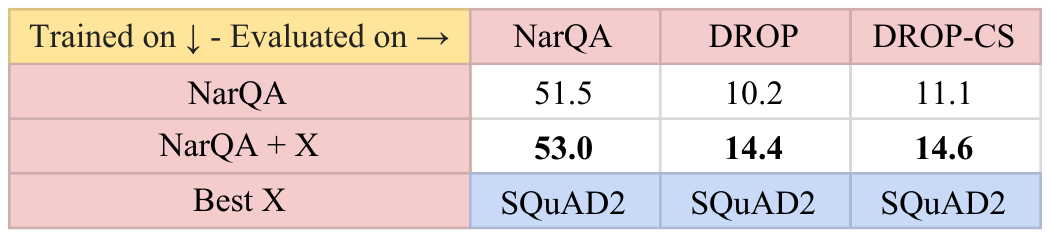}
     \caption{
        % Pairwise mixing of formats: fine-tuning with additional datasets (of different formats) boosts performance. All the results use T5 with a ``small'' size.
        % \daniel{BoolQ table should have 1 digit after the pointer.}
        Pilot study showing that out-of-format training can help improve performance. Each table compares training on just the anchor dataset (e.g., BoolQ in the top-left table) with training also on an out-of-format dataset denoted `X'. Evaluation is on the anchor dataset as well as unseen datasets of that format. The last row identifies the out-of-format dataset that helped most on each evaluation dataset.
        All results are based on the ``small'' size T5 model.
        Color denotes QA format (see Table~\ref{fig:datasets:properties}). 
        % \hanna{This figure has a way too many datasets for MC; I suggest cut some of the MC datasets (ARC easy, QASK and CQA all just showing squad11 is the best) and then we can fit table 1 and 2 in one row.}
     }
     \label{tab:pairwise_table}
 \end{table*}

%%%%%%%%%%%%%%%%%
%%%%%%%%%%%%%%%%%
\subsection{Evaluation Metrics for Textual Output}

We 
% train and 
evaluate 
% \ashish{correct, or is training objective different??} \daniel{they're different/independent; good catch! } 
each dataset using the metric used most often for it in prior work. For the EX format, it's the F1 score of the extracted span relative to the gold label. For the AB format, we use ROUGE-L metric~\cite{lin2006information,min-etal-2019-discrete,nishida-etal-2019-multi}. 
For NatQA we use the exact-match metric, following \citet{min2020ambigqa}. 
%For the MC format, we match the generated the text with the closest answer candidate (by token overlap) and measure how often this is the correct answer.
For the MC format, we match the generated text with the closest answer candidate based token overlap and compute the accuracy.
% \sewon{Rewrote this - please check.}
% \oyvind{Maybe define better "closest" here? For yes/no, would an answer "yes it is" be counted as "yes" or as credit 0 because it's not exactly "yes" or "no"? There's a fuzzy boundary here when it's not defined by a task-specific prefix exactly what class of answer we're looking for, an important issue when combining increasing numbers of tasks like this...}
For the YN format, we follow \citet{clark-etal-2019-boolq} to measure if the generated output matches the correct `yes' or `no' label. In rare cases where the output is longer than one word (e.g., `yes it is'), we check if it contains the correct label but not the incorrect one.\footnote{The evaluation code \added{is available at the URL in Footnote~\ref{footnote:code}.}}

\section{Pilot Study: Can Out-of-Format Training Help?}
\label{subsec:pair}
% \tushar{alt. name?: Impact of out-of-format datasets or Proof of concept: Pairwise Mixing}

We first answer the question: \emph{Is the broad idea of benefiting from out-of-format training even viable?} For instance, is our intuition correct that an MC dataset can, in practice, benefit from training on an EX dataset? Before discussing our main experimental results, we briefly report on a pilot study that assesses the following basic question: Given a training set $T^i_1$ (the \emph{anchor} dataset) of QA format $F_i$, is there an out-of-format training set $T^j_1$ of format $F_j$ such that training jointly on $T^i_1 \cup T^j_1$ improves performance relative to training only on $T^i_1$? To this end, we evaluate both on the matching evaluation set $E^i_1$ as well as on `unseen' data $E^i_2, E^i_3, \ldots$ of the same format.

% The first question we address is whether there is value in mixing datasets with different formats. We evaluated this by adding one dataset of a different format to four different datasets (one for each format).
%Towards this goal, we mix pairs of different formats and evaluate them. 

The results are summarized in Table~\ref{tab:pairwise_table}. The two rows in each individual table correspond to training on $T^i_1$ (the \emph{anchor} dataset) and on $T^i_1 \cup X$, where $X$ is an out-of-format dataset corresponding to $T^j_1$ above. The columns represent various evaluation sets of format $F_i$. For each column, `$X = \ldots$' at the very bottom indicates the out-of-format dataset $X$ that was the most helpful in improving performance on the evaluation set in that column.\footnote{Appendix~\ref{appendix:subsec:pairwise} reports extended results, including the performance with various choices of $X$.}

Consider the case of the anchor set $T^i_1$ being BoolQ and the evaluation set being NP-BoolQ, both of format YN. Here, including out-of-format training data $X{=}$SQuAD2 boosts performance from 51\% to as much as 59\%. The gain may be less in other cases, but across all anchor and evaluation datasets, we generally observe that there is at least one out-of-format training set whose inclusion improves performance.

This pilot study thus provides a proof of concept that out-of-format training can indeed help a QA model in nearly every case. Of course, this study only shows the existence of such an out-of-format dataset, rather than provide a single unified model. Nevertheless, it helps identify \emph{representative training sets} from each format that were most helpful. As alluded to earlier, we used this empirical data to guide which training sets to include when building \unifiedQA in Section~\ref{subsec:unifiedQA}.

\begin{figure}[ht]
    \centering
    \includegraphics[scale=0.50,trim=4.65cm 4.6cm 2cm 2.5cm]{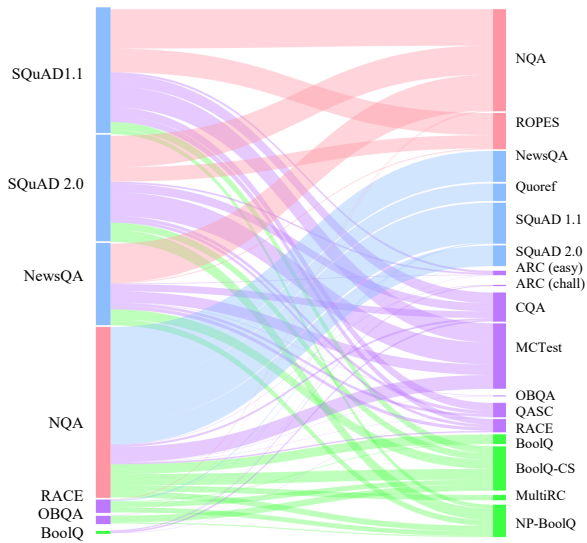}
    \caption{Bipartite graph showing the value of various datasets. The datasets on the left were used for training and on the right for evaluation. The wider the edge from a dataset $\ell$ (on the left) to a dataset $r$ (on the right), the higher the contribution of adding the out-of-format dataset $\ell$ to the training set of questions in $r$'s format. 
    % Stronger edges show bigger contribution.
    }
    \label{fig:bipartite-fig}
\end{figure}

\added{
The experimental results from this case study are summarized in the aggregated plot shown in Fig.~\ref{fig:bipartite-fig}. 
In this bipartite graph, the datasets used for training are on the left hand side and the evaluation datasets are on the right hand side. The weight of each edge $w(\ell, r)$ indicates the contribution of a dataset $\ell$ when used for training jointly with an anchor dataset $d$, when evaluated on $r$ ($d$ and $r$ have the same format.) Specifically, \\
%edge weights are computed as:   
% \begin{align*}
% w(\ell, r) = avg_{d} \Big[ & \text{score} \big(\text{training on } \ell \cup d, \text{ eval. on } r\big) \\ 
% - &  \text{score} \big(\text{ training on } d; \text{eval. on } r\big) \Big].  
% \end{align*}
%\begin{align*}
\indent $w(\ell, r) = avg_{d} \Big[ S\big(\ell \cup d; r\big) -  S\big(d; r\big) \Big],$ \\ 
%\end{align*}
where $S(d, r)$ is the score achieved on $r$ after training on $d$. Since we only focus on \emph{gains} from out-of-format training, we drop the edges that are negative or between two datasets of the same format.
%. Additionally, we exclude all $(\ell, r)$ pairs that are from the same format.

As expected, there are strong connections between the AB and EX datasets in Fig.~\ref{fig:bipartite-fig} since their definitions are quite similar. Apart from the edge weight, the overall width of a dataset $\ell$ on the left also depicts how much it contributes to out-of-format datasets. E.g., NQA (NarrativeQA) is the most helpful dataset and even helps multiple formats. Similarly our extractive datasets (SQuAD11.1, SQuAD 2, and NewsQA) are also relatively more helpful. %These large datasets help a lot as they can be possibly help models learn basic comprehension skills. 
While large datasets generally appear to help, RACE, another large-scale dataset, doesn't help that much. The least helpful dataset in the mix is BoolQ which focuses on yes/no questions. 

In a similar vein, the wider the dataset on the right hand side, the more it can be benefit from out-of-format datasets. Among these beneficiary datasets, all four formats are equally represented.  
}

%%%%%%%%%%%%%%%%%%%%%%%%
\section{Experimental Results}
\label{sec:experiments}
% In this section we build a \emph{single} model that works on \emph{multiple} formats. 
% Not surprisingly, in our union mixture we need representatives from each format. 
% In particular, we create a \unifiedQA that is trained on the union of the following datasets: SQuAD 1.1, SQuAD 2.0, NQA, Regents, ARC, MCTest, RACE, OBQA, BoolQ. 
% For the rest of the draft, we use the largest available T5 model (11B), unless otherwise specified.

We now discuss our main experimental results, evaluating \unifiedQA on seed datasets (used for training the system) as well as unseen datasets. 

\begin{table*}[tb]
    \centering
    \includegraphics[scale=0.66,trim=1.95cm 15.5cm 2cm 2cm]{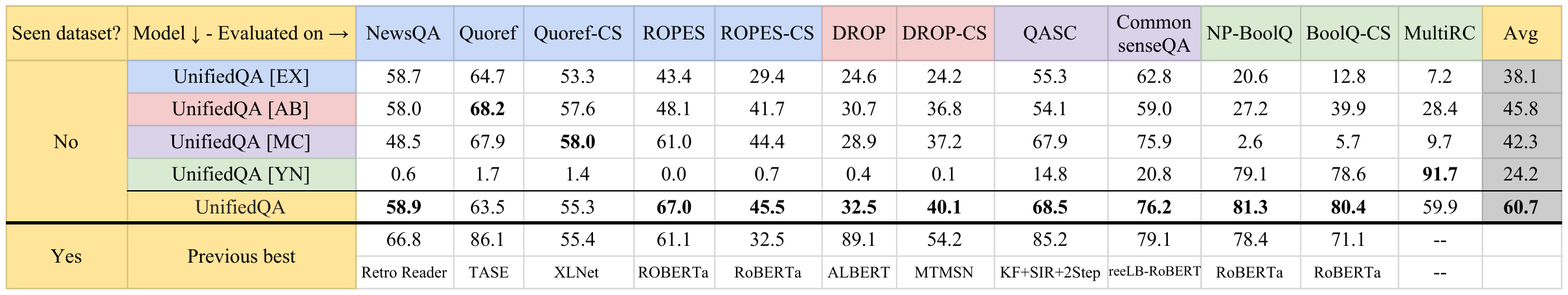}
    \caption{
        % \daniel{for Quoref, AB should be bold. }
        % \daniel{Why is UnifiedQA [MC] better than UnifiedQA [EX]: check the numbers. }
        Generalization to unseen datasets: Multi-format training (\unifiedQA) often outperforms models trained the same way but solely on other in-format datasets (e.g., \unifiedQA[EX], which is trained on all extractive training sets of \unifiedQA. When averaged across all evaluation datasets (last column), \unifiedQA shows strong generalization performance across all formats. Notably, the ``Previous best'' models (last row) were trained on the target dataset's training data, but are even then outperformed by UnifiedQA (which has \underline{never seen these datasets} during training) on the YN tasks.
    }
    \label{tab:generalization}
\end{table*}

\subsection{\unifiedQABig vs.\ 8 Dedicated Models}
\label{subsec:union:vs:single:dataset}

Is \unifiedQA, a single pre-trained multi-format QA system, as good as dedicated systems trained for individual datasets? 
We emphasize that the answer to this question is not as simple as it may seem, since earlier works have observed that a system addressing multiple tasks often \emph{underperforms} a focused system~\cite{raffel2019exploring}.

\begin{figure}[t]
    \centering
    \includegraphics[scale=0.41,trim=0.2cm 1.1cm 0cm 0.5cm]{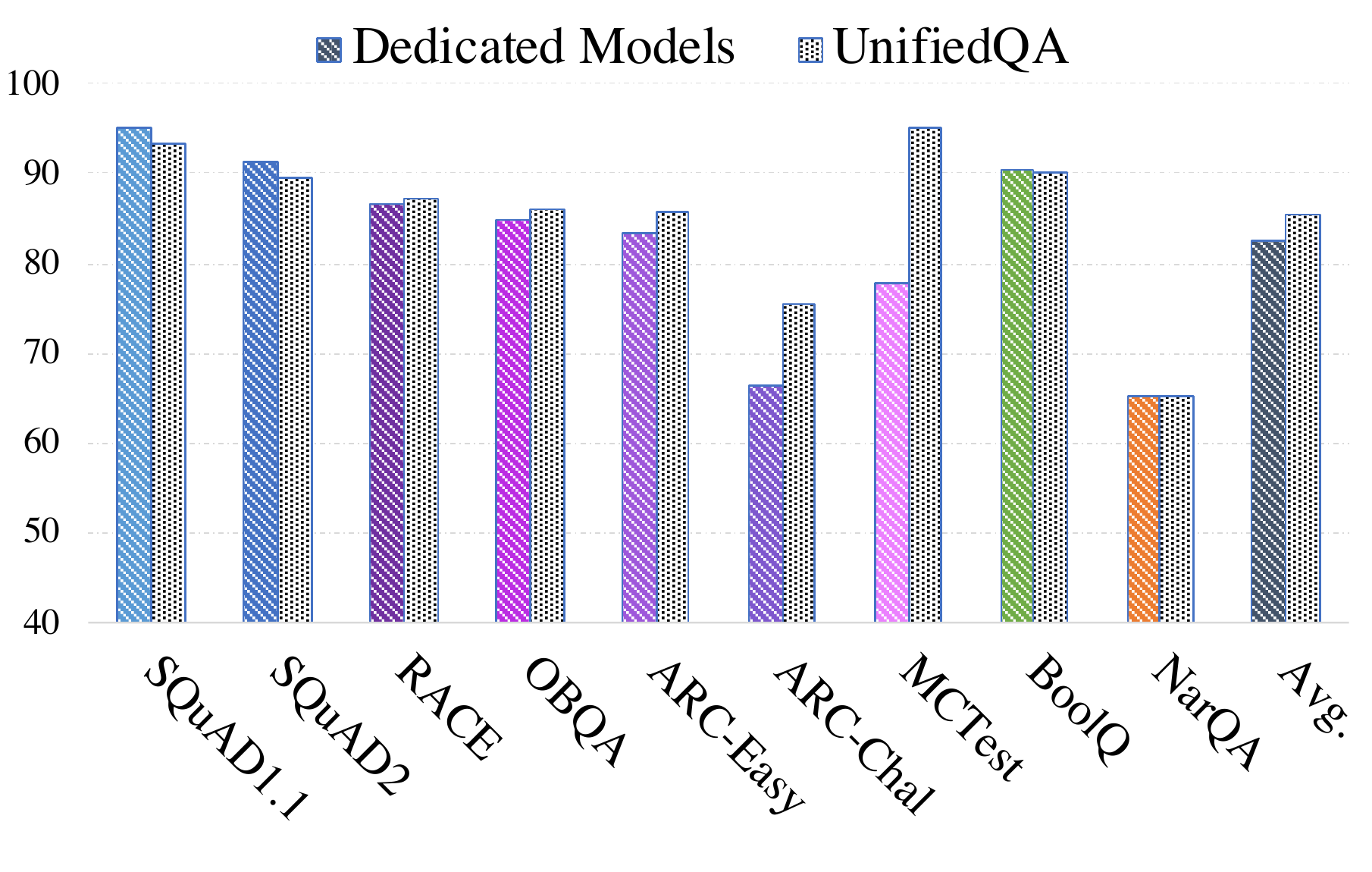}
    \caption{
        \unifiedQA is on-par with, and often outperforms, 9 different equally-sized T5-based systems tailored to individual datasets.
        The figure contains separate models for each of the two subsets of the ARC and Regents datasets. \
        % hanna{This figure does not have the average bar chart... we refer to it in the paper.}
        % \ashish{should say something to reconcile these 11 columns with the earlier mentions of 9 datasets. Something like this? Here we train two separate models each for subsets of ARC and Regents datasets. Not sure. But it's confusing at the moment.}
    }
    \label{fig:union:vs:single:dataset}
\end{figure}

Fig.~\ref{fig:union:vs:single:dataset} summarizes the results of the relevant experiment. 
The gray bars belong to \unifiedQA\ (a single system for multiple datasets of different formats). 
The colored bars are different T5-based systems tailored to individual datasets (a different system for each dataset).
The results show that \unifiedQA performs almost as good as
% best single dataset experts
individual T5 models targeted to each dataset. 
In some cases \unifiedQA performs even better than the single-dataset experts (e.g., on OBQA or NQA). 
On average (last column) \unifiedQA clearly outperforms the ensemble of dataset/format-specific systems. 
\unifiedQA thus offers flexibility across multiple QA formats while compromising almost nothing compared to dataset-specific experts. 
% \unifiedQA performs as good as dataset-specific experts while working on several different formats. 

\begin{table*}[t]
    \centering
    \includegraphics[scale=0.66, trim=1.9cm 14.55cm 1.8cm 1.85cm,clip=true]{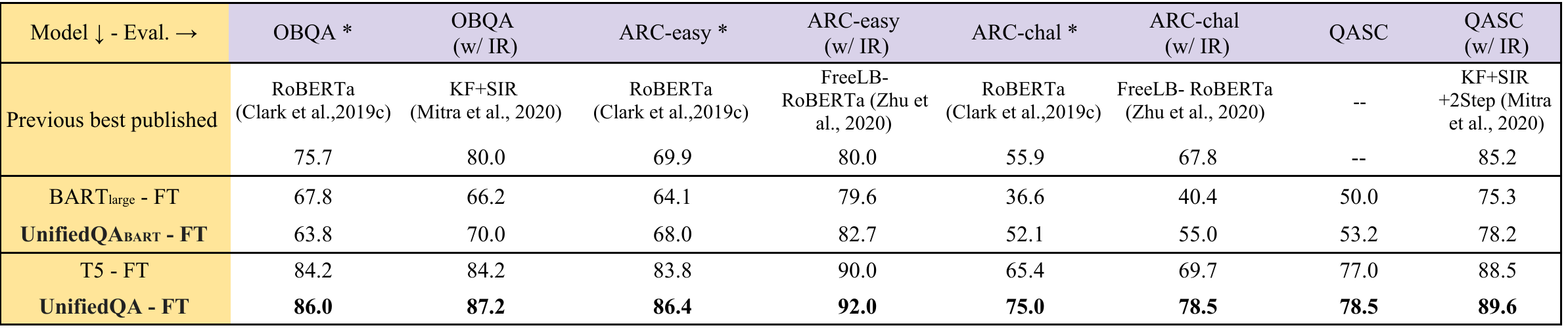}
    \includegraphics[scale=0.67, trim=1.0cm 10.30cm 3cm 6.9cm,clip=true]{figures/sota10.pdf}
    \caption{
        Fine-tuning \unifiedQA (last row) results in new state-of-the-art performance on 11 datasets. Further, it consistently improves upon fine-tuned T5 (2nd last row) by a margin ranging from 1\% for CommonsenseQA (CQA) to as much as 13\% for ARC-challenge. `(w/ IR)' denotes relevant information is retrieved and appended as context sentences in the input encoding. Datasets marked with * are used in \unifiedQA's original training. 
        % For datasets marked with $\bullet$ we use \unifiedQA/T5 of size 3B; otherwise the models are 11B. 
        % \hanna{Very nitpicky: but it is a bit weird that we have names of systems in the 2nd row -- makes it a bit hard for direct comparison to the numbers below} \ashish{agreed. better?}
        % \ashish{having the `best' row somewhere is the middle is non-standard. should we switch to: previous best, T5 FT, UQA, UQA FT?  Or perhaps previous best, UQA, T5 FT, UQA FT?  Either way, better to end with UQA FT.} 
        % \hanna{color code the datasets.}\tushar{It is. They are all MCQ datasets.}
        % \sewon{The red one should be NatQA instead of NarQA?}
        % \sewon{Sorry to ask for change, but I feel like the name SpanSeqGen may be misleading; `DPR+BART' might be a better name. Would you update the table? (Already changed it in Sec 6.3.}
        % \sewon{On NatQ, maybe 49.3 should be bold?}
        % \sewon{If we run out of space, adding the citation for prev. SOTA as a caption and removing them in the main text would also work, IMHO..}
    }
    \label{tab:fine-tuning}
\end{table*}

%%%%%%%%%%%%%%%%%%%%%%%%
\subsection{Generalization to Unseen Datasets}
\label{subsec:generalization}

We now explore whether \unifiedQA generalizes well to other, unseen datasets. 
Table~\ref{tab:generalization} summarizes the results of experiments where we evaluate various models on datasets that are not used to train them. It compares \unifiedQA\ (training on multiple formats) with training on various datasets of a \emph{single} format (e.g., \unifiedQA[EX], built by training the model on only extractive datasets). 

% \hanna{need to explain what unifeidqa[ex] means}

The first few rows of the table show T5 models trained for individual formats, followed by \unifiedQA.
For completeness, we include the highest previous scores for each dataset; one must be careful when reading these numbers as the best previous numbers follow the fully \emph{supervised} protocol (for NewsQA~\cite{zhang2020retrospective}, Quoref~\cite{segal2019simple}, DROP~\cite{lan2019albert}, ROPES~\cite{lin-etal-2019-reasoning}, QASC~\cite{khot2019qasc}, CommonsenseQA~\cite{Zhu2020FreeLB} and x-CS datasets~\cite{gardner2020evaluating}.) 
% \sewon{
% It seems like the dataset paper is cited for DROP, instead of SOTA paper (ALBERT, \citet{lan2019albert})?
% }

We make three key observations: 
(1) On average (last column), \unifiedQA shows much stronger generalization across a wide range of datasets. 
(2) on 9 (out of 12) datasets, \unifiedQA shows a better generalization than any single-format expert. 
For example, while the system is trained on multiple-choice questions with 4 candidate answers, it works quite well on datasets with more than 4 candidate answers (QASC and CommonsenseQA have has 8 and 5 candidate answers per question, respectively).
(3) Single-format experts are better at generalization only when the source and target datasets are very similar (for instance SQuAD and Quoref). 
% In this section we compare \unifiedQA with best previous system for each dataset. 

\begin{table*}[t]
    \centering
    \includegraphics[scale=0.785,trim=3.8cm 13.8cm 2cm 1.4cm]{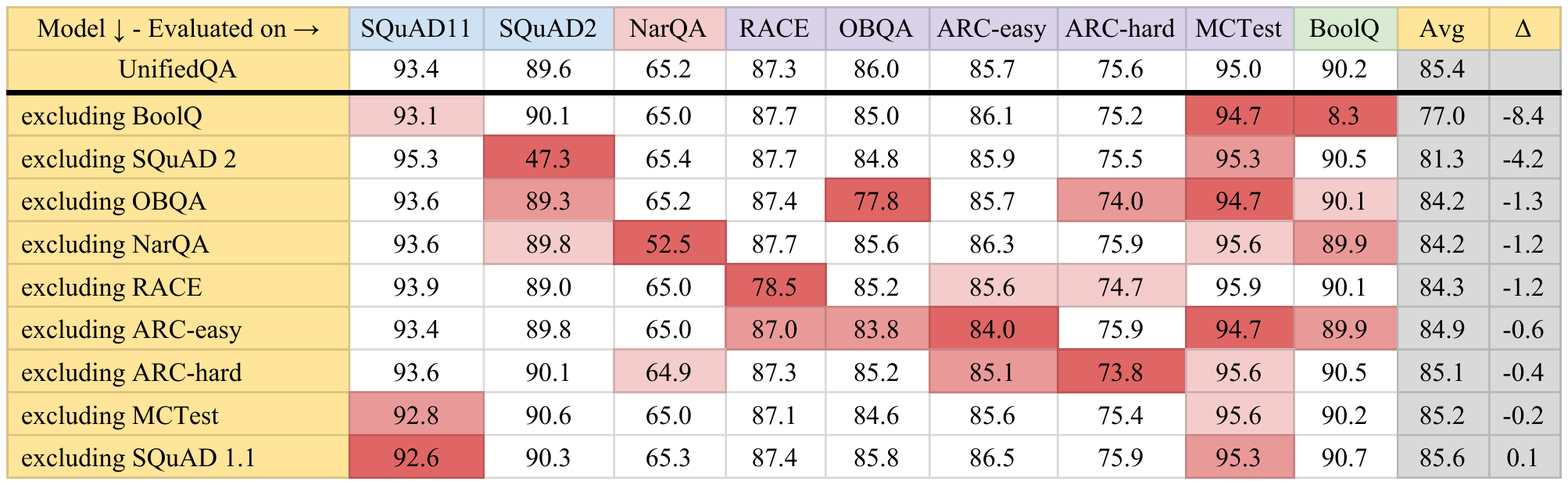}
    \caption{
        The results of a leave-one-out ablation. The first row indicates the performance of \unifiedQA on each dataset it was trained on. The rest of the rows exclude one dataset at a time. The rows are sorted based on the last column: the dataset with biggest contribution appear first. The \redtext{red highlights} indicate the top 3 performance drops for each column.
        % \sewon{NQA to be NarQA?}
    }
    \label{tab:leave:one:out}
\end{table*}

%%%%%%%%%%%%%%%%%%%%%%%%
% \subsection{Fine-tuning Staring with \unifiedQABig}
% \subsection{Achieving SOTA w/ \unifiedQABig}
\subsection{State-of-the-Art via Simple Fine-tuning}
\label{subsec:sota}
Fine-tuning of pre-trained language models has become the standard paradigm for building dataset-specific stat-of-the-art systems~\cite{devlin-etal-2019-bert,liu2019roberta}. 
The question we address here is: when it comes to QA, is there a value in using \unifiedQA as a starting point for fine-tuning, as opposed to a vanilla language model that has not seen other QA datasets before? 

To address this question, we fine-tune each of \unifiedQA, T5, and BART on several datasets by selecting the best check point on the dev set, and evaluating on the test set. 
Table~\ref{tab:fine-tuning} summarizes the results of the experiments. 
The table shows two variants: \unifiedQATfive\ and \unifiedQABART. 
All results are based on the 11B version of T5. %versions of T5 models. 
% wherever possible, except for datasets indicated with $^\bullet$.
% \footnote{Due to memory limitations imposed by the leaderboards.} 

% , build based on  T5$_\text{11B}$ and BART$_\text{large}$ architectures, respectively.  
% \sewon{ Also curious what you think about names like \unifiedQA$_\text{T5}$/\unifiedQA$_\text{BART}$ or \unifiedQA-\textsc{t5}/\unifiedQA-\textsc{bart}, or change T5(11B) and BART(large) to T5$_\text{11B}$, BART$_\text{large}$. Just felt like there are too many parentheses overall in the paper. (Would love to be responsible for changing if you want.)
% } \hanna{+1; I like these index-style presentation vs. parenthesis}
% The two last rows of the table show the performance \unifiedQA and T5, both fine-tuned for the target task. %The fine-tuning process involves selection of the best checkpoint on the dev set and evaluation on the test set. 

The columns indicate the evaluation on the test set corresponding to the data that was used for training. 
For each dataset, the first line of the table reports the best previously published work.
% The first row of the table ``used in the union'' indicate whether the dataset was observed 
For several MC datasets that do not come with evidence  paragraphs, we include two variants: one where we use them as-is and another that uses paragraphs fetched via an Information Retrieval (IR) system as additional evidence, indicated with ``w/ IR'' tags. We use the same IR sentences as used by the baselines: Aristo corpus for ARC and OBQA datasets~\cite{clark2019f}, and 2-step IR for QASC~\cite{khot2019qasc}. 
For NatQA, following \cite{min2020ambigqa}, we use the DPR retrieval engine~\cite{karpukhin2020dense} to augment each question with additional paragraphs.

% \tushar{Changed. also not sure what you used for obqa.} \daniel{used thr IR paragraphs Oyvind game me.}
%\tushar{For QASC, are you using the 2-step IR?} \daniel{I am using the IR paragraphs you game me.}. 
% \daniel{clarify w/ IR and w/ IR.}
% \daniel{clarify the row of "used in the union".}

\nocite{lan2019albert}
\nocite{clark2019f}
\nocite{Mitra2020HowAK}
\nocite{banerjee2020knowledge}
\nocite{Zhu2020FreeLB}
\nocite{min2020ambigqa}

\begin{comment}
Additionally, we show the best published scores on each dataset: 
ALBERT~\cite{lan2019albert} (on RACE), 
RoBERTa~\cite{clark2019f} (on OBQA, ARC, Winogrande/WG, PIQA, SIQA~\cite{Mitra2020HowAK}), 
KF+SIR~\cite{banerjee2020knowledge} (on OBQA and QASC), 
FreeLB+RoBERTa~\cite{Zhu2020FreeLB} (on ARC-easy and CommonsenseQA), and 
DPR+BART~\cite{min2020ambigqa}.
\end{comment}
% \tushar{Just a list of citations seems odd. Should we just add it to the table or leave this to the Appendix?}
% \daniel{What is `odd' about assigning the credits? `Oddness' is in the eye of the beholder! :) 
% Putting them in the table would need space. 
% I don't like putting them in the appendix as it is too detached from where it belongs. 
% I am not sure-concerned about space; that being said, I am open to other suggestions to make it more readable. 
% }
% \sewon{I agree we should cite prev. SOTA in the main paper not in Appendix, but also think it is a bit repetitive with a row in Table 5 and hurts the flow of this section.. How about citing them as a caption? Or, maybe keep the model names but remove data names, as which model corresponds to SOTA of which model is already indicated in the table?
% } \hanna{+1; I like the idea of citing them as the caption of the table.}\tushar{Nothing odd about giving credit but the flow is odd :) Footnote or caption sounds good. Adding a row to the tables, shouldnt take much more space and avoids adding coreferences}

We see that fine-tuning on \unifiedQA consistently dominates fine-tuning on T5 and BART, respectively. It also dominates the best previous scores on the datasets. 
Intuitively, since \unifiedQA has seen different formats, it should be positioned to achieve higher scores after a little fine-tuning, compared to fine-tuning on a vanilla T5 or BART model.
% \tushar{Maybe worth mentioning that you also showed that the effectiveness of cross-format training is not limited to T5 only.}
% \sewon{+1. I guess it's obvious from the table but still seems nice to mention it explicitly.}
This could be especially effective when a user has limited training data for a target QA task (also shown in Appendix~\ref{appendix:subsec:winogrande}.) 
This also highlights that the effectiveness of cross-format training is not limited only to T5, but is rather a general trend for text-to-text architectures. 

% For completeness, the final row of shows best previously published numbers for each dataset (SQuAD 1.1 and BoolQ from~\cite{raffel2019exploring}, SQuAD 2 and RACE from~\cite{lan2019albert}, NQA from~\cite{nishida-etal-2019-multi} and OBQA from~\cite{clark2019f}). 

% \hanna{probably worth adding the point that you report the best previously pulished results, but I think you are still better than most others in the leaderboard. It is worth mentioning that and refer to the corresponding leaderboards.}

% \hanna{Question: Shall we mention something about the model size? I am not sure!}

%%%%%%%%%%%%%
\subsection{Ablation: Training Set Contributions}
\label{subsec:leave-one-out}

We now perform a leave-one-out experiment to better understand the contribution of each seed dataset to \unifiedQA. 
We take the system from \S\ref{subsec:unifiedQA} and assess how strong the model is when individual seed training datasets are dropped from the union. 
The result of this experiment is summarized in Table~\ref{tab:leave:one:out}. It compares the performance of full \unifiedQA (the first row) with ablated variants that exclude one seed dataset at a time. 
The rows are sorted based on the last column: datasets with higher contributions appear first.
% \oyvind{Not sure if it's worth mentioning as people have moved on from 1.1, but the 95.3 SQuAD 1.1 result without the confounding no-answer 2.0 training is basically SOTA.}
% \daniel{my SQuAD numbers are dev tho, since I can't evaluate on their official test files.}

Looking at first few rows of the table, BoolQ, SQuAD 2.0, OBQA, NarQA are the top four contributing datasets, each with a different format. 
% \sewon{Changed NQA to NarQA.}
SQuAD 1.1 has the least importance, presumably because it is mostly covered by SQuAD 2.0. 
% \sewon{Actually, do we have any specific reason for including SQuAD v1.1 although it is a subset of SQuAD 2.0?} \daniel{nope, it became clear that SD1 is redundant after doing this ablation. Btw, SD1 is NOT a strict subset of SD2. }

This study suggests that in order to build an effective unified QA system, it suffices to have a relatively small set of datasets as long as the set includes representatives from each format.

\added{
%%%%%%%%%%%%%%%%%%%%%%%%
\section{Discussion}

The key motivation for this work is the observation that nearly all prior efforts on QA research were limited to the boundaries defined by narrow \emph{formats}. A \emph{format-specific} design would not generalize across QA datasets with slightly different definitions (e.g., a model built for SQuAD would not work for RACE). 
Additionally, such a design would prevent us from benefiting from the labeled data available in other formats. 
We challenge this view by advocating for approaches that combine seemingly different datasets. 
We believe that developing QA systems targeted to a specific format is a conceptual barrier for progress in the field. 
% In summary, our unifiedQA setting enables integrating multi-format questions across various domains with different sizes.

% It is encouraging that \unifiedQA{} has already been studied and shown to be quite strong in terms of its generalization to other domains in the recent literature~\cite{hendrycks2020measuring}. 

% \hanna{vote to remove this paragraph} %We believe, our work is aligned with the gradual trend of earlier years about moving towards the most general definition of tasks; e.g., the transition towards ``Universal Dependencies'' in the Parsing literature~\cite{nivre2016universal}.

\paragraph{Factors affecting generalization.}
Format is not the only factor affecting generalization across datasets. We additionally studied the value of other factors including \emph{dataset size} and \emph{domain} (vocabulary,  topic, and style) in improving generalization.  
We observed that larger datasets often help with generalization, but not always (\S\ref{subsec:pair}); e.g., RACE or OBQA show similar benefits (Fig.~\ref{fig:bipartite-fig}), even though RACE is much larger than OBQA. 
% \hanna{instead, you can say: we observe similar gains on xxx when we augment NewsQA or OBQA to ???}
We observed a similar phenomenon with domain: similar domains help with transfer, but that is not always the case. For example,  
while BoolQ questions, similar to SQuAD, are accompanied with Wiki paragraphs, they barely benefit each other. 
% Wikipedia-based questions helped to Commonsense datasets, although the domains are dissimilar. \hanna{can you give another example? like a dataset that do not helps another dataset the domains are similar?} 
Overall, the factors affecting generalization are not well-understood, leaving room for future investigations.

\paragraph{Unifying QA formats and text-to-text models.}
While \unifiedQA is built based using existing text-to-text models~\cite{gpt2,raffel2019exploring}, we emphasize that the choice of tasks for multi-task learning plays a crucial role in achieving successful results. Previous studies \cite{raffel2019exploring} did \emph{not} observe gains when mixing {\it tasks} that are very different.  The key intuition is that a more coherent choice of {\it tasks} is more likely to succeed. 
Further, focusing on a coherent space of QA tasks/formats allows us to simplify the input by not requiring ``prefixes'' to explicitly define tasks/formats.
}

%%%%%%%%%%%%%%%%%%%%%%%%
\section{Conclusion}

The question-answering community has fruitfully explored the design of strong models, but while staying within the boundaries of individual QA formats.
% This is the first work to cross the boundaries of different formats, to the best our knowledge.  
We argued that such boundaries are artificial and can even limit the performance of systems, because the desired reasoning abilities being taught and probed are not tied to specific formats. Training data in one format should, in principle, help QA systems perform better even on questions in another format.

With this intuition in mind, we presented \unifiedQA, a single pre-trained QA system based on the text-to-text paradigm, seeking to bring unification across four common QA formats. 
% This is through several controlled experiments: 
% (1) there transfer when mixing pairs of datasets that could help systems trained multiple formats (\ref{subsec:pair}), 
We showed that even with its simple multi-format training methodology, \unifiedQA achieves performance on par with 8 dataset-specific expert models (\S\ref{subsec:union:vs:single:dataset}), while also generalizing well to many unseen datasets of seen formats (\S\ref{subsec:generalization}). 
At the same time, we demonstrated that \unifiedQA is a strong starting point for building QA systems: it can achieve state-of-the-art performance by simply fine-tuning on target datasets (\ref{subsec:sota}).  

We hope this effort will inspire a future line of work in the QA and NLP communities, moving towards more general and broader system designs. 
% Needless to mention that this work does not cover all the possible QA formats.  
We leave extensions of \unifiedQA to other formats such as to direct-answer questions~\cite{Roberts2020HowMK} as a promising avenue for future work.
% \sewon{What is the context for citing Natural Questions? Or is it intended to be \citet{Roberts2020HowMK}?}

% in this study. 
% as we deemed them significantly different from the rest of our formats. 
% We did try it in our initial experiments: turns out that it almost never helps. However, in order to have a more general system one should also include it in their union. 
% We leave these as potential avenues to study in the future work. 

% \oyvind{Not sure if/where to mention a caveat with combining datasets: Incidental leakage between train/test data in different datasets. This is already somewhat of an issue with pretrained LMs, e.g., if they have read a lot of Wikipedia they might be overly good at Wikipedia-centered tasks like SQuAD}
% \daniel{this sounds like a much broader issue that our focus (i.e., cross the format-boundaries). While I agree that it's an important issue and deserves attention, I doubt that this is right place.}

% \daniel{
% include a reference to the sota results and mention our recommended practice: if you want to build a qa dataset, especially if you have a limited data, fine-tune with \unifiedQA. 
% }

%%%%%%%%%%%%%%%%%%%%%%%%
\subsection*{Acknowledgments}
The authors would like to thank Collin Raffel, Adam Roberts, and Nicholas Lourie for their help with the T5 framework and for providing feedback on an earlier version of this work. 
The authors would like to acknowledge grants by ONR N00014-18-1-2826 and  DARPA N66001-19-2-403, and gifts from the Sloan Foundation and the Allen Institute for AI. Moreover, the authors would like to thank members of the Allen Institute for AI, UW-NLP, and the H2Lab at the University of Washington for their valuable feedback and comments.
TPU machines for conducting experiments were provided by Google.

% \begin{small}
\bibliography{additional}
\bibliographystyle{acl_natbib}
% \end{small}

%%%%
% \newpage
\clearpage

\appendix
\onecolumn

\section{Appendices}
\label{sec:appendix}

\subsection{Datasets: Details}
\label{appendix:sec:datasets}
% \hanna{I suggest to distinguish between "seed datasets" and "evaluated datasets"}

% \hanna{if no space, this section can be very much summarized.}
% \ashish{+1. we are spending 1 page on 4.1. Let's move this as-is to the appendix, and have a really succinct version here. Cover dataset names + citations, and only point out unusual / important pieces of info (e.g., that DROP requires math skills (worth pointing out as it's different)).}
We evaluate our \unifiedQA on 19 existing datasets that target various formats, as well as various complex linguistic phenomena. 
Table~\ref{fig:datasets:properties} shows different properties for our datasets (whether it comes with a paragraph, 
% domain of the questions/paragraphs,
whether the paragraph explicitly contains the answer, 
whether there are candidate-answers as part of the input, etc.) 
Most importantly, they are grouped into several formats/categories described below. 
% : \textbf{EX} (extractive questions), \textbf{MC} (multiple-choice), \textbf{YN} (yes-no questions), \textbf{AB} (abstractive questions). 
% \nick{I don't think you need to list the categories here, you can just say "described below" or something and let them read the paragraphs.}
% From each format, I have 1 or 2 datasets selected to train on. For example, in the YN category, I am training on BoolQ. 
Table~\ref{tab:statitstics} gives summary statistics of these datasets. 

\paragraph{Extractive QA (EX).} All the datasets in this format require models to extract the answer to a given question as a substring from a context paragraph. 
SQuAD 1.1~\cite{rajpurkar-etal-2016-squad} contains questions about Wikipedia paragraphs. 
A later version of this dataset, SQuAD 2~\cite{rajpurkar-etal-2018-know}, includes unanswerable questions which empirically makes the task much harder. 
For our evaluation, we use the development sets of SQuAD 1.1 and SQuAD 2. 
NewsQA~\cite{trischler-etal-2017-newsqa} dataset focuses on paraphrased questions with predicate-argument structure understanding collected from news articles from CNN/DailyMail articles.
Quoref~\cite{dasigi-etal-2019-quoref} contains questions that require coreference resolution in Wikipedia articles and can even have disjoint spans as answers. %extracted from Wikipedia articles. 
% multi-span answers from Wikipedia articles that centered around pronouns in the context. 
ROPES~\cite{lin-etal-2019-reasoning} centers around situation understanding, where the model must under the causes and effects implicit in the given situation. 

\paragraph{Abstractive QA (AB).}
All the datasets in this format require models to produce answers that are often not mere substrings of the given context paragraph. 
NarrativeQA~\cite{kocisky-etal-2018-narrativeqa} focuses on understanding various events that happen in a given movie plot, based on summaries of their movie adaptations from various web resources. 
Many of the answers do not have high overlap with the context. 
DROP~\cite{dua-etal-2019-drop} contains questions that involve rudimentary mathematical skills (such as counting, addition, subtraction, maximum, minimum, etc.) and questions query multiple parts of the paragraph. 
The answer can be either a number or a date that can be inferred from the paragraph, or several spans from the context paragraph.
Finally, we use an open-domain version of NaturalQuestions~\cite{kwiatkowski-etal-2019-natural} where the paragraph that was used for creating the question is eliminated, and only the questions with short answers up to five tokens are taken.
% \sewon{
%     I updated it to add that only short answers are included.
% } 
% \daniel{thanks!}
Instead, we follow~\cite{min2020ambigqa} to use a DPR retrieval~\cite{karpukhin2020dense} engine to augment each question with an additional context paragraph. We call this dataset NatQA.

\paragraph{Multiple-choice QA (MC).}
All the datasets in this format contain questions that come with candidate answers. 
MCTest~\cite{richardson-etal-2013-mctest} contains questions about simple, fictional stories. 
RACE~\cite{lai-etal-2017-race} is a challenging set of English comprehension multiple choice exams given in Chinese middle and high schools. 
OpenBookQA~\cite{mihaylov-etal-2018-suit}, ARC \cite{clark2018think,clark2016combining}, 
% Regents Science Exams \cite{clark2016combining}, 
QASC~\cite{khot2019qasc} are different MC tests focusing on elementary/high school-style science exams. 
We use several othern datasets that are often framed as commonsense reasoning benchmarks: 
CommonsenseQA~\cite{talmor-etal-2019-commonsenseqa} is geared towards activity/concept questions, PIQA~\cite{bisk2019piqa} addresses physical interaction reasoning, 
SIQA~\cite{sap2019socialiqa} contains question that require social reasoning (motivations, reactions, event orders) and finally Winogrande~\cite{sakaguchi2019winogrande} which a benchmark for hard pronoun resolution problems~\cite{Levesque2011TheWS,peng2015solving}.

Other than  MCTest and RACE, the rest of the datasets do not come with accompanying paragraphs. On such datasets, occasionally a retrieval system is used to supplement each question with a relevant retrieved context paragraph. For most of this the work, we keep the questions as is with no additional retrieval (unless otherwise mentioned), except in \S\ref{subsec:sota} where we use IR to get numbers comparable to earlier work. 
One other variability among these datasets is their number of candidate answers. While many datasets have four candidates (see Figure~\ref{fig:datasets:properties}), others have more. Later, in \S\ref{subsec:generalization} we will see that our approach generalizes to datasets with different number of candidates, even if it's not seen during training. 
% \daniel{rework this: the point I wanna make here is generalization across different candidate-answer sizes, while not explicitly encoding the number of the candidates.}

\paragraph{Yes/No QA (YN).}
All the datasets in this format contain questions that could be responded with yes/no answers. One can think of these as multiple-choice questions with 2 candidates; however, they're usually treated differently. Several examples we use are BoolQ~\cite{clark-etal-2019-boolq} and a version of this dataset with natural-perturbations, BoolQ-NP~\cite{khashabi2020naturalperturbations}, 
the subset of MultiRC~\cite{khashabi-etal-2018-looking} that have binary(yes/no) answers. 

\paragraph{Contrast-sets.}
Additionally, we use \emph{contrast-sets}~\cite{gardner2020evaluating} for several of our datasets (denoted with ``CS''): BoolQ-CS, ROPES-CS, Quoref-CS, DROP-CS. 
These evaluation sets are expert-generated perturbations that deviate from the patterns common in the original dataset.

\subsection{Details on the experiments: }
\label{appendix:subsec:hyperparams}
Below is several details on the experiments: 

\begin{itemize}
    \item \underline{Models: } we use two text-to-text frameworks: T5 and BART. 
    \item \underline{Model sizes: } Most of the experiments are done on T5(11B) which has 11 billion parameters. We also report experiments with BART (large) with 440 million parameters. 
    \item \underline{Input/output size: } For all experiments, we use token-limits of size 512 and 100 for inputs and outputs sequences, respectively.  
    \item \underline{\# of iterations for pretraining on the seed datasets (\S\ref{sec:multiformat-training}):} All models are trained for $100k$ steps on the seed datasets.
    \item \underline{Learning rates:} we use 1e-3 and 1e-5, for T5 and BART, following the original works on each framework. 
    \item \underline{Batch sizes:} We use batches of 8 and 120, for the T5 (11B) and BART models, respectively. 
    % \sewon{Update 128$\xrightarrow{}120$.}
    \item \underline{Infrastructure: } In the experiments, we use v3-8 TPUs for T5 models, and eight 32GB GPUs for BART models. 
    \item \underline{Time spent to build \unifiedQA:} pretraining \unifiedQA\ approximately takes about 36 and 55 hours, on T5(11B) and BART models, respectively. 
    \item \underline{Finetuning on datasets (\S\ref{subsec:sota}):} the only hyperparameter we iterated over is the training steps. Each model was fine-tuned for 60$k$ steps and checkpoints were saved every 2$k$ steps. The model with the highest score on the dev set is our selected model.
\end{itemize}

% Number of parameters in each model 
% Corresponding validation performance for each reported test result 
% Explanation of evaluation metrics used, with links to code  
% Hyperparameter configurations for best-performing models (*)
% Number of hyperparameter search trials (*)
% The method of choosing hyperparameter values (e.g., uniform sampling, manual tuning, etc.) and the criterion used to select among them (e.g., accuracy) (*)

\clearpage

\subsection{\unifiedQABig: Different Sizes}
\label{appendix:subsec:unified:qa:sizes}

 For completeness we're also showing the scores of \unifiedQA of different sizes on each dataset. For these systems each row is a single system. 
%  , unlike the other rows where we have a single system per column. 
%  \oyvind{This is not part of the table anymore, but maybe worth a table in appendix?}

\begin{table}[h]
    \centering
    \includegraphics[scale=0.66,trim=1.8cm 15.7cm 0cm 2.2cm]{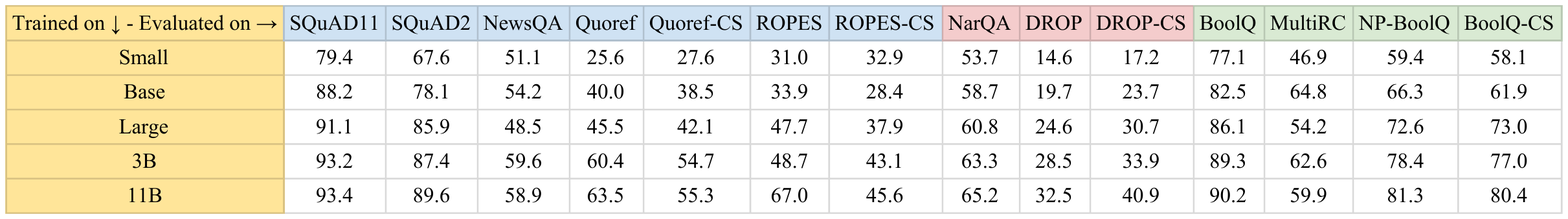}  
    \includegraphics[scale=0.66,trim=1.8cm 16.2cm 0cm 2.2cm]{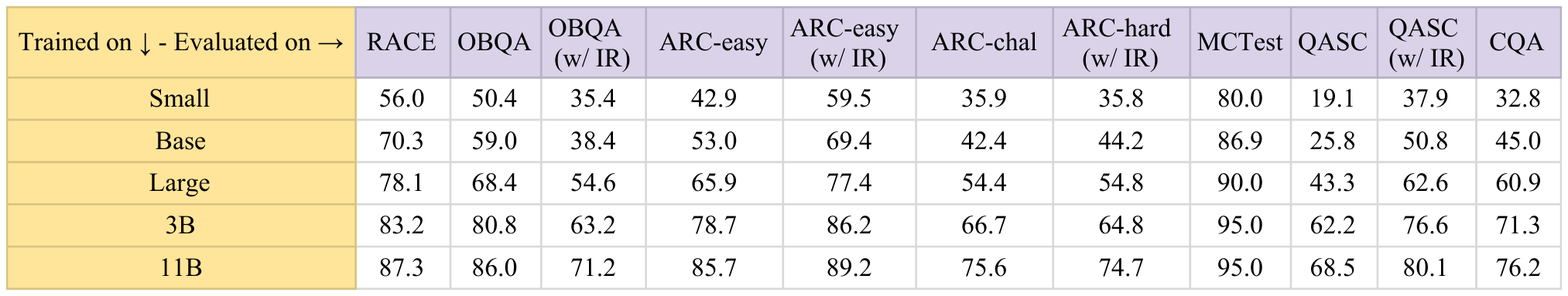}  
    \caption{\unifiedQA of different sizes on our datasets. 
    % \daniel{we have repeated OBQA's in the 2nd table} 
    }
    \label{tab:unifiedqa:different:model:sizes}
\end{table}

\subsection{Comparison with the Dedicated Models: extended results}
\label{appendix:subsec:union:vs:single:dataset}

Here we summarize an extension of the results in \S\ref{subsec:union:vs:single:dataset}.
Table~\ref{tab:appendix:union:vs:single:dataset} summarizes the results of the relevant experiment.  In the top portion of the table we have evaluations of T5 model fine-tuned for individual datasets, followed by \unifiedQA. As it can be observed from the table, \unifiedQA performs almost as good as the best single dataset experts. In some cases \unifiedQA performs even better than than the single-dataset experts (e.g., on OBQA or NQA.)  On average (last column) \unifiedQA is doing much better dataset/format-specific systems. 
In conclusion, \unifiedQA offers flexibility across multiple QA formats while compromising almost nothing compared to dataset-specific experts.

 \begin{table*}[h]
    \centering
    \includegraphics[scale=0.665,trim=2.1cm 13.9cm 2cm 1.5cm, clip=false]{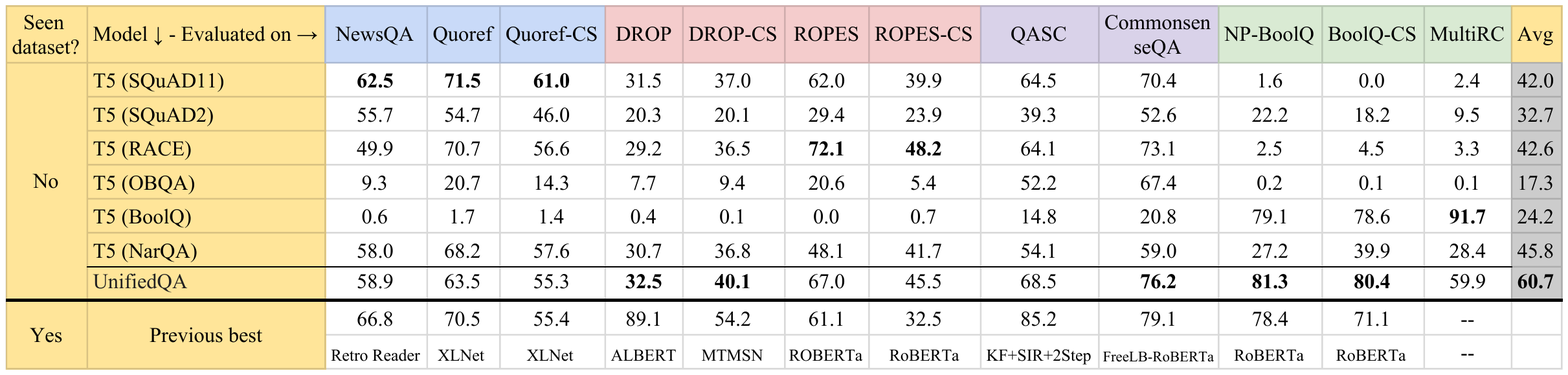}
    \caption{
        \unifiedQA is on-par with systems tailored to individual datasets (the diagonal cells vs the last row) while functioning across a wide range of datasets (the last column). 
        % \ashish{off-diagonal entries here are an unfair comparison, as they don't get to see the training data whereas \unifiedQA does. Can't defend reading those in a meaningful way.} 
        % \ashish{A little too many rows/columns here. 11 as opposed to the 9 we have mentioned elsewhere, and strongly dominated by purple (MC) datasets. How about combining both ARCs together, and both Regents together, to get 9 total?}
    }
    \label{tab:appendix:union:vs:single:dataset}
\end{table*}

\subsection{Pairwise Mixing: extended results}
\label{appendix:subsec:pairwise}
Here we summarize an extension of the results in \S\ref{subsec:pair}.
The question addressed here is whether there is value in mixing datasets with different formats. 
We evaluated this by adding one dataset of a different format to four different datasets (one for each format).
The results are summarized in Table~\ref{tab:appendix:pairwise_table}.  
The goal of each sub-table is to measure the \emph{within-format} generalization one can gain via \emph{out-of-format} training.  
Each sub-table has an \emph{anchor} dataset, indicated in the first column. For example in the first table the anchor dataset is SQuAD. Rows of the table: 
Each table combines datasets of other formats with the anchor dataset (e.g., SQuAD + RACE, etc). The columns of the sub-tables contain evaluations on the dataset with the same format as the anchor dataset. For example, on the first table, the evaluation is done on SQuAD 1.1/2.0, NewsQA, Quoref which have the same format as SQuaD 1.1, the anchor dataset. 
The results show that one can achieve gains for question-answering in a certain format by incorporating resources in other formats. 
In the first two sub-tables, we see that NarQA (AB) and OBQA (MC) help a SQuAD models generalize better to other EX datasets. 
In the third table where the anchor dataset is NQA (AB), EX datasets help a NQA model generalize better to other AB datasets. 
In the 4th/5th subtable, EX and AB datasets help a RACE/OBQA (MC) models generalize better to other MC datasets. 
Similarly, in the final sub-table, MC dataset helps improve the scores on a YN datasets.

 \begin{table*}[!htbp]
     \centering
     \includegraphics[scale=0.64,trim=3cm 13cm 2cm 1.5cm, clip=false]{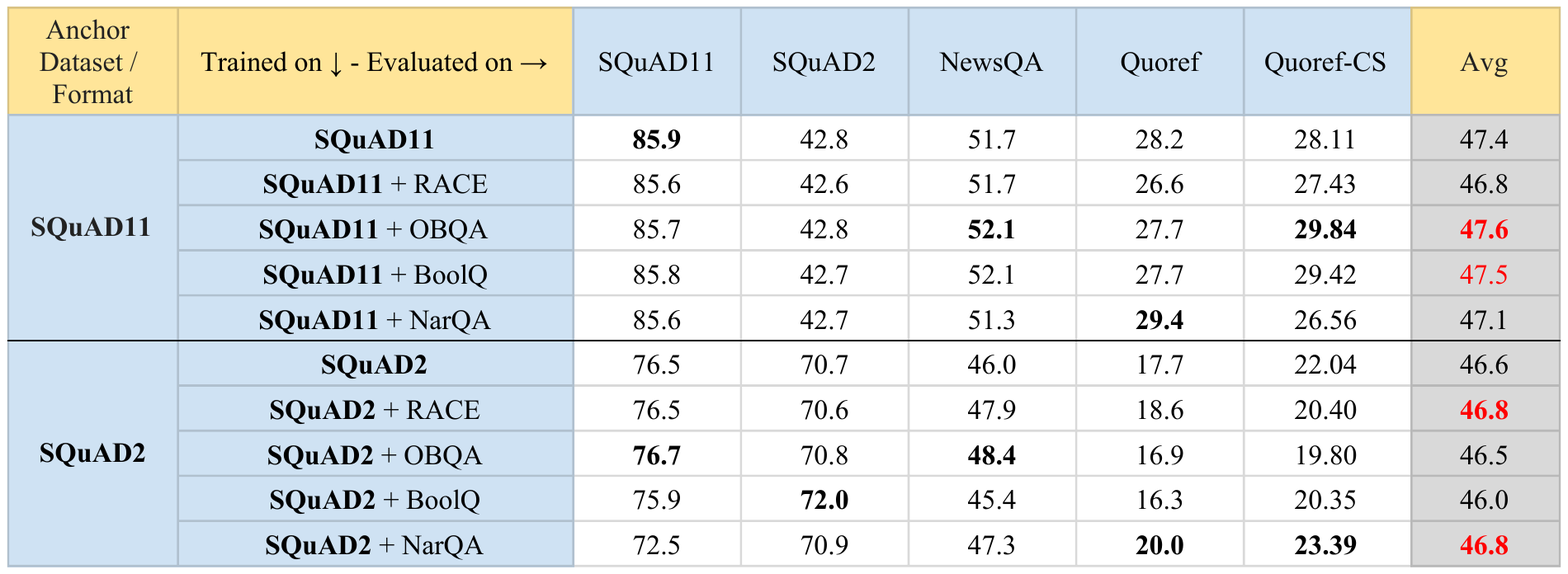}
     \includegraphics[scale=0.64,trim=3cm 14.7cm 2cm 1.5cm, clip=false]{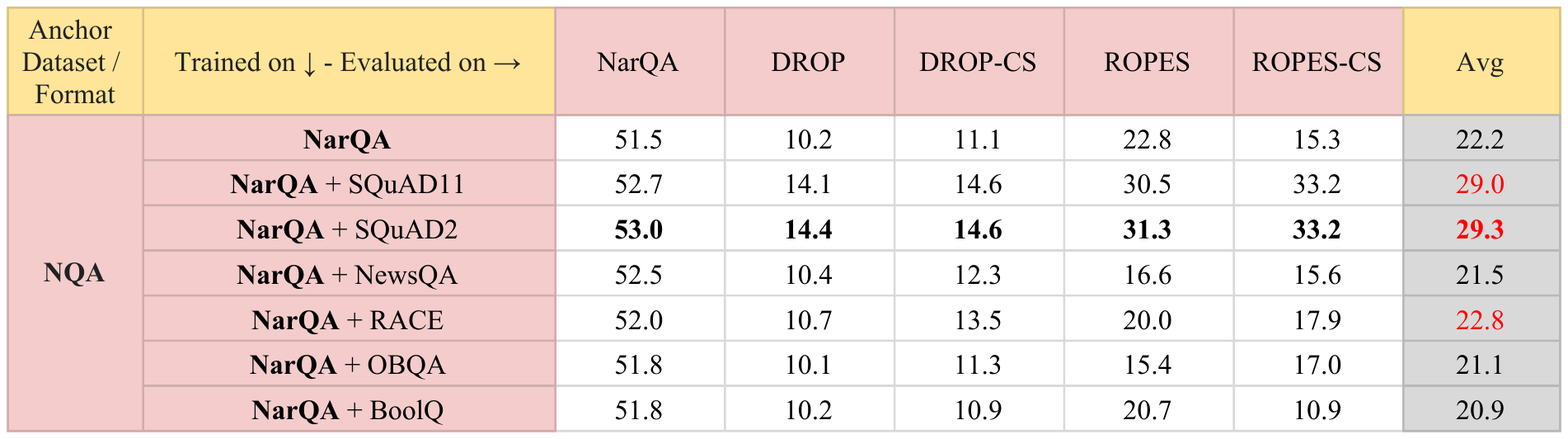}
     \includegraphics[scale=0.64,trim=3cm 12.5cm 2cm 1.5cm, clip=false]{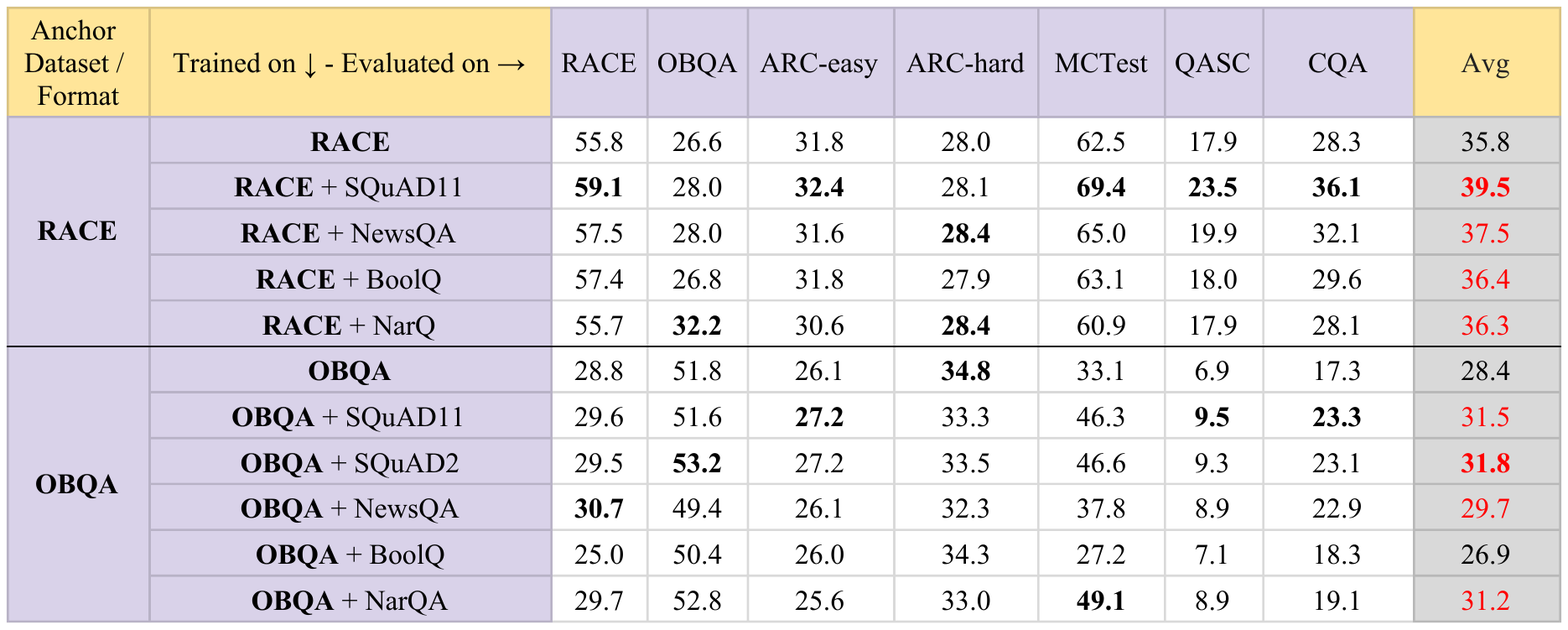}
     \includegraphics[scale=0.64,trim=3cm 14.7cm 2cm 1.5cm, clip=false]{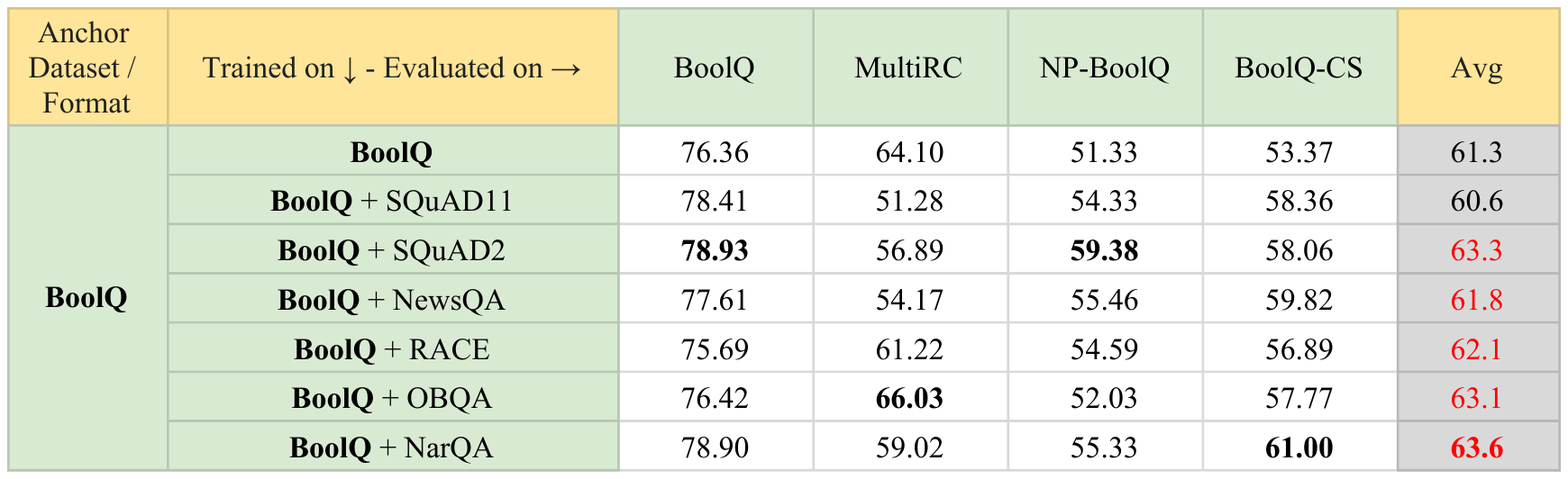}
     \caption{Pairwise mixing of formats: mixing with QA of datasets with different formats helps. }
     \label{tab:appendix:pairwise_table}
 \end{table*}
 
 \clearpage
 
\subsection{Extended Results of Fine-tuning on Winogrande}
\label{appendix:subsec:winogrande}

Here we provide extended result for the Winogrande dataset. The results are summarized in Table~\ref{tab:winograd:table}. 
The table include results of fine-tuning \unifiedQATfive\ and \unifiedQABART, as well as fine-tuning of the vanilla language models, T5 and BART. 
As it can be observed, on this dataset, fine-tuning  \unifiedQA gives stronger results when the size of the training data is limited.  
With respect to the overall metric AUC, \unifiedQA has a slight edge over fine-tuning the vanilla language models. 

 \begin{table}[ht]
     \centering
    %  \begin{framed}
     \includegraphics[scale=0.8,trim=4cm 16cm 0cm 2cm]{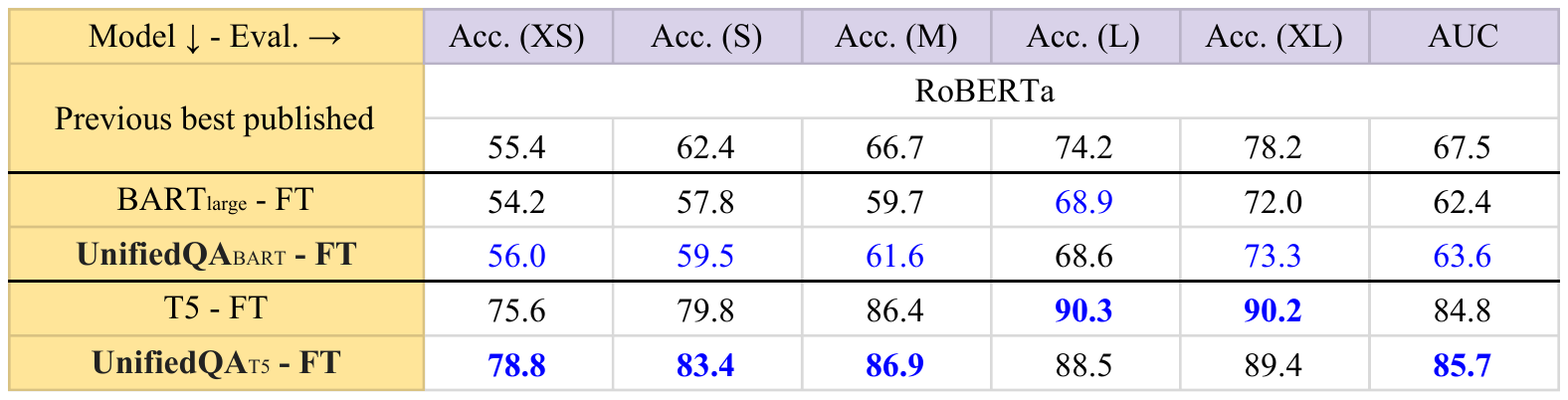}
    %  \end{framed}
     \caption{Extended results on the Winogrande dataset}
     \label{tab:winograd:table}
 \end{table}

\end{document}